\PassOptionsToPackage{dvipsnames, table}{xcolor}
\documentclass{article} 
\usepackage{iclr2024_conference,times}

\usepackage{amsmath,amsfonts,bm}

\def\eqref#1{equation~\ref{#1}}

\def\1{\bm{1}}

\DeclareMathAlphabet{\mathsfit}{\encodingdefault}{\sfdefault}{m}{sl}
\SetMathAlphabet{\mathsfit}{bold}{\encodingdefault}{\sfdefault}{bx}{n}

\usepackage[table]{xcolor}
\usepackage{hyperref}
\usepackage{url}
\usepackage{array}
\usepackage{graphicx}
\usepackage{booktabs}
\usepackage{appendix}
\usepackage{tabularx}
\usepackage{multicol}
\usepackage{multirow}
\usepackage[utf8]{inputenc}
\usepackage[T1]{fontenc}
\usepackage{fontawesome}  
\usepackage{tcolorbox}
\usepackage{array}       
\usepackage{ragged2e}    
\usepackage{pifont} 
\usepackage{caption}
\usepackage{xspace}

\definecolor{mydarkblue}{rgb}{0,0.08,0.45}
\definecolor{citeblue}{HTML}{3b86d9}

\hypersetup{
    colorlinks=true,
    linkcolor=mydarkblue,
    filecolor=blue,      
    urlcolor=mydarkblue,
    citecolor=citeblue,
}

\newcommand{\cmark}{\textcolor{green}{\ding{51}}} 
\newcommand{\xmark}{\textcolor{red}{\ding{55}}}   

\title{OpenCoder: The Open Cookbook for Top-Tier Code Large Language Models}

\newcommand{\binf}{{\raisebox{-0.1em}{\includegraphics[height=0.9em]{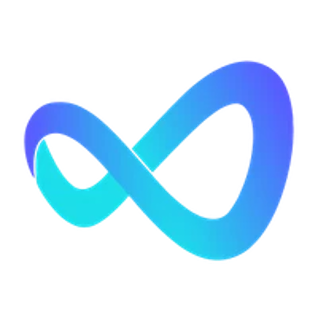}}}}
\newcommand{\bmap}{{\raisebox{-0.1em}{\includegraphics[height=0.9em]{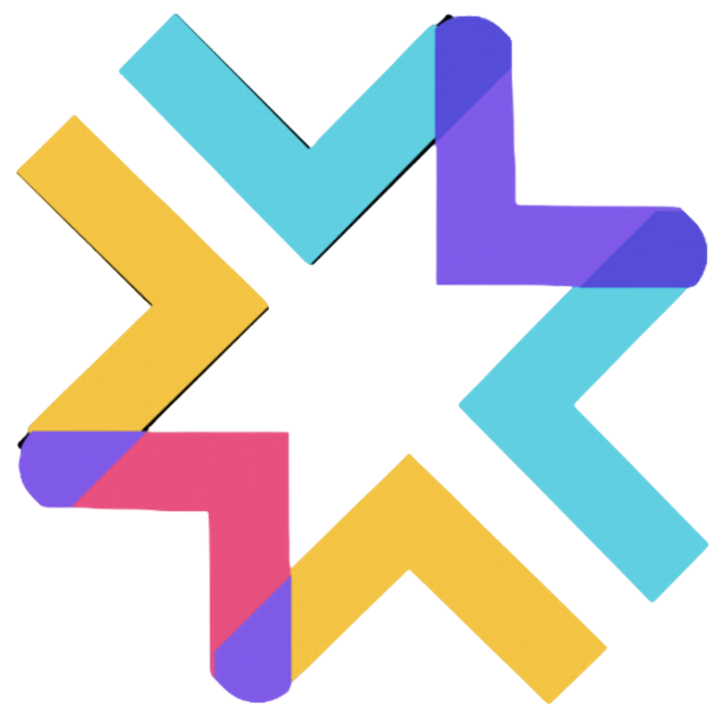}}}}
\newcommand{\bnju}{{\raisebox{-0.1em}{\includegraphics[height=0.9em]{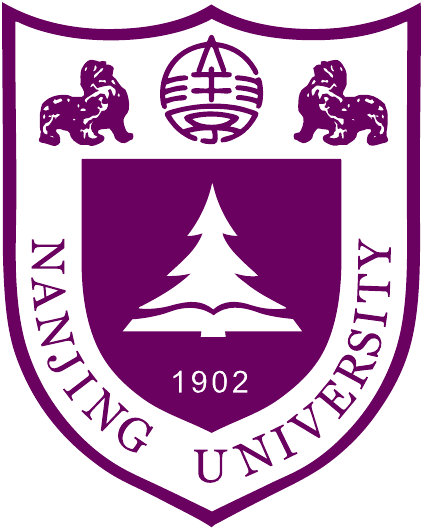}}}}

\newcommand{\homepage}{\raisebox{-4.0pt}{\includegraphics[height=1.8em]{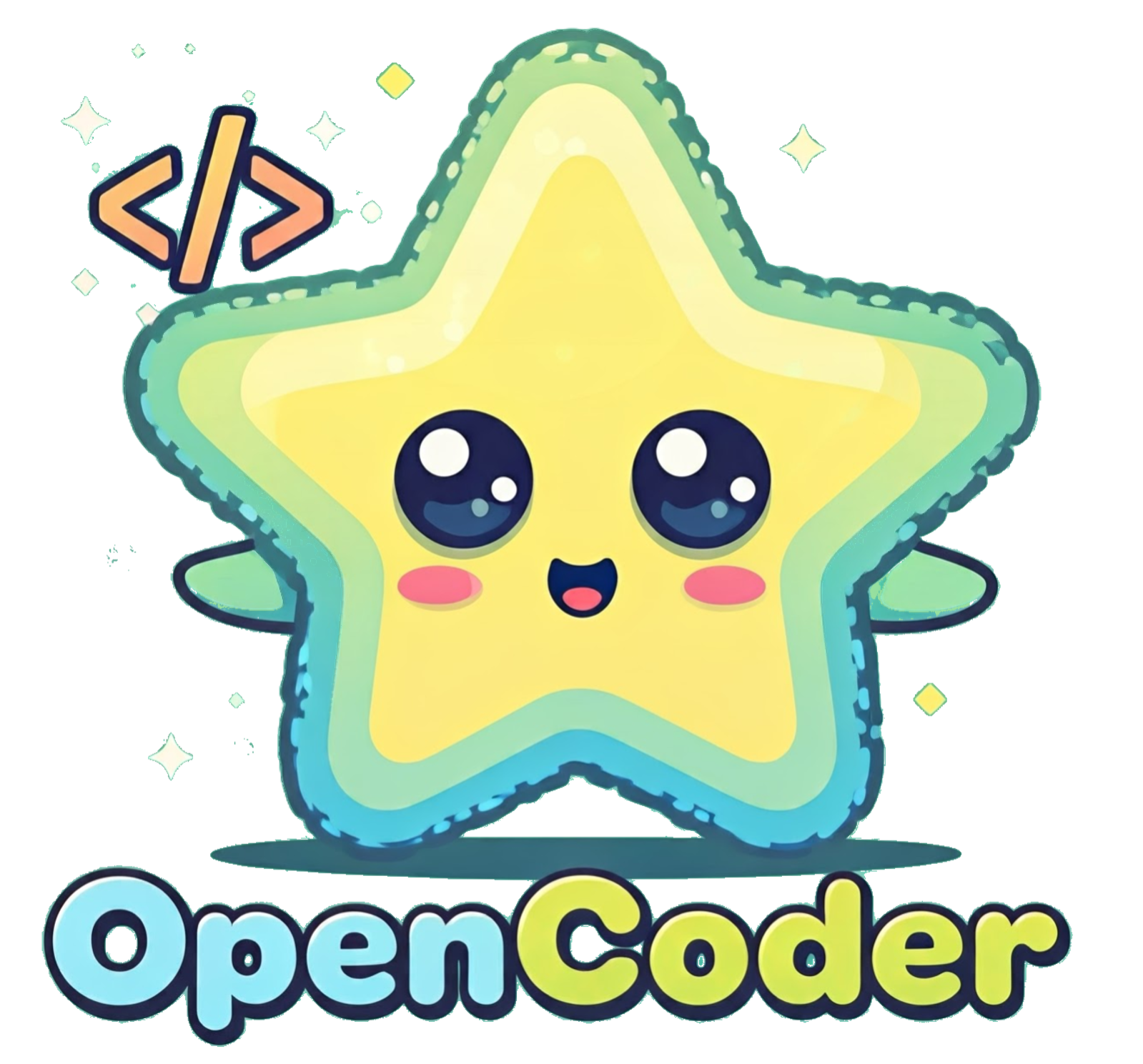}}\xspace}

\author{
    {\textbf{\!\!Siming Huang}}\thanks{The first two authors contributed equally to this work. Work done during the internships of Siming Huang and Tianhao Cheng at INF. $^\dagger$ Correspondence to Wei Chu (chuwei@inftech.ai) and Zili Wang (ziliwang.do@gmail.com).}\;\,$^\binf$  \ \
    {\textbf{Tianhao Cheng}}$^*$$^\binf$ \ \
    {\textbf{J.K. Liu}}  \ \ 
    {\textbf{Jiaran Hao}}$^\binf$ \ \ \\
    {\textbf{Liuyihan Song}}$^\binf$  \ \ 
    {\textbf{Yang Xu}}$^\binf$ \ \
    {\textbf{J. Yang}}$^\bmap$ \ \ 
    {\textbf{Jiaheng Liu}}$^\bnju$ \ \
    {\textbf{Chenchen Zhang}}$^\bmap$ \ \ \\
    {\textbf{Linzheng Chai}}$^\bmap$ \ \ 
    {\textbf{Ruifeng Yuan}}\;\; \ \
    {\textbf{Zhaoxiang Zhang}}\;\; \ \
    {\textbf{Jie Fu}}\;\; \ \ 
    {\textbf{Qian Liu}}\;\; \ \ \\
    {\textbf{Ge Zhang}}$^\bmap$ \ \
    {\textbf{Zili Wang}}$^\dagger$ $^\binf$ \ \
    {\textbf{Yuan Qi}}$^\binf$ \ \ 
    {\textbf{Yinghui Xu}}$^\binf$ \ \ 
    {\textbf{Wei Chu}}$^\dagger$ $^\binf$ \ \
    \\ \\
    $^\binf${\rm{INF}} \ \
    $^\bmap${\rm{M-A-P}} \ \
    $^\bnju${\rm{NJU}} \ \ \\
    \\
    {\homepage Home Page: \url{https://opencoder-llm.github.io}}
}

\iclrfinalcopy 
\begin{document}

\maketitle

\vspace{-0.6cm}
\begin{abstract}
Large language models (LLMs) for code have become indispensable in various domains, including code generation, reasoning tasks and agent systems.
While open-access code LLMs are increasingly approaching the performance levels of proprietary models, high-quality code LLMs suitable for rigorous scientific investigation, particularly those with reproducible data processing pipelines and transparent training protocols, remain limited.
The scarcity is due to various challenges, including resource constraints, ethical considerations, and the competitive advantages of keeping models advanced.
To address the gap, we introduce OpenCoder, a top-tier code LLM that not only achieves performance comparable to leading models but also serves as an ``open cookbook'' for the research community.
Unlike most prior efforts, we release not only model weights and inference code, but also the reproducible training data, complete data processing pipeline, rigorous experimental ablation results, and detailed training protocols for open scientific research.
Through this comprehensive release, we identify the key ingredients for building a top-tier code LLM: (1) code optimized heuristic rules for data cleaning and methods for data deduplication, (2) recall of text corpus related to code and (3) high-quality synthetic data in both annealing and supervised fine-tuning stages.
By offering this level of openness, we aim to broaden access to all aspects of a top-tier code LLM, with OpenCoder serving as both a powerful model and an open foundation to accelerate research, and enable reproducible advancements in code AI.
\end{abstract}
\vspace{-0.4cm}

\begin{figure}[htbp]
    \centering
    \includegraphics[width=1.0\textwidth]{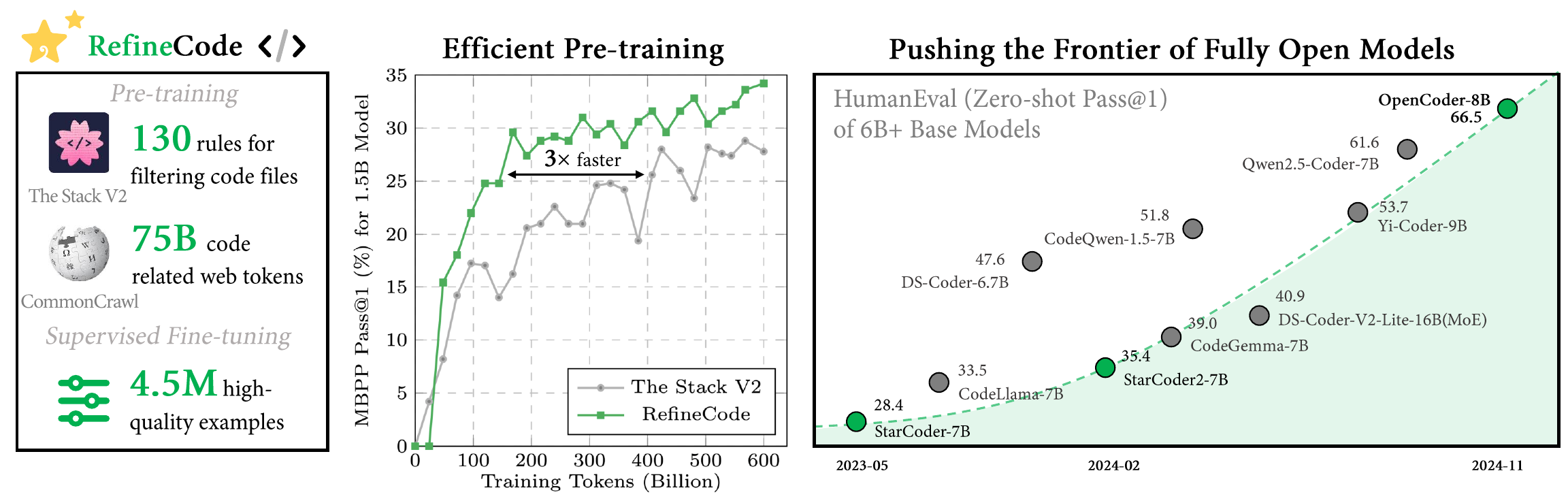}
    \caption{OpenCoder surpasses all previous fully open models (i.e., with open model weights and reproducible datasets) and other open-access models (i.e., with open model weights only) at the 6B+ parameter scale, pushing the frontier of fully open models to new heights.}
    \label{fig:enter-label}
\end{figure}

\newpage
\tableofcontents
\newpage

\section{Introduction}


Large Language Models (LLMs) have achieved significant success in various domains~\citep{wang2023rolellm,Que2024DCPTLD,liu20242,liu2024ddk,wu2024conceptmath}, particularly in code-related tasks, revolutionizing the current paradigm of software development~\citep{qian2024chatdev,wang2024opendevin}. Code-specific LLMs have emerged as a critical area within LLM research, with tools such as ChatGPT, Copilot, and Cursor reshaping the workflows of developers. Despite this, the performance of open-source LLMs focused on code~\citep{li2023starcoder,taocrystal,lozhkov2024starcoder,zhang2024map} still falls short compared to state-of-the-art LLMs~\citep{hui2024qwen2,zhu2024deepseek}, largely because these leading models keep their training datasets—an essential factor in LLM development—proprietary. This lack of transparency limits the broader research community’s ability to establish strong baselines and gain deeper insights into the workings of top-tier code LLMs.

To remedy the gap, we set forth three primary goals by releasing \textbf{OpenCoder} and its development material: 
(1) Firstly, we aim to provide scholars with a meticulously curated and fully transparent strong baseline code LLM for research on mechanical interpretability and the data distribution of code LLMs.
(2) Secondly, we intend to conduct in-depth investigations into the pretraining and instruction data curation pipeline for the development of stronger code LLMs. 
(3) Thirdly, by enabling a detailed review of the development of the models, we hope to unlock more diverse customized solutions based on transparent code LLM.
Through OpenCoder, we strive to stimulate and accelerate the growth of the open-source code LLM community.


Our comprehensive set of controlled experiments highlights key design choices for data curation for top-tier code LLMs in different training stages:
(1) During the pretraining phase, the importance of data cleaning is highlighted~\citep{zhou2024programming}, emphasizing the removal of non-informative data such as pure hexadecimal code and excessively short code snippets that do not contribute to the learning process.
(2) The impact of deduplication is significant, with file-level deduplication proving to be more effective than repository-level deduplication by maintaining data diversity and enhancing model performance on downstream tasks~\citep{li2023starcoder}.
(3) The influence of GitHub stars is also examined, revealing that filtering data based on Github star count can possibly reduce data diversity and affect the overall data distribution, contributing to a suboptimal result~\citep{allal2023santacoder}. 
(4) In the annealing phase, the use of high-quality data is crucial for further enhancing the model's capabilities, indicating that data quality is more important than quantity in the later stages of model training. 
(5) Finally, during the instruction tuning phase, a two-stage instruction tuning strategy is shown to be effective, allowing the model to acquire broad capabilities initially and then refine them with code-specific tasks, resulting in improved performance on both theoretical and practical coding tasks. These five key points underscore the importance of data quality, diversity, and targeted enhancement strategies in developing a high-performing code generation model like OpenCoder.

This work introduces the OpenCoder, a completely open-source Code LLM, built on the transparent data process pipeline and reproducible dataset. As shown in Table \ref{table:coder_comparison}, We provide the open cookbook to build a code LLM from scratch by providing the data cleaning pipeline, reproducible pretraining dataset, large-scale SFT Corpus, and intermediate checkpoints. OpenCoder, through its meticulous data processing and advanced training methods, has surpassed expectations by achieving top-tier results on multiple code LLM evaluation benchmarks. The introduction of the open cookbook of code LLM is designed to push forward the field of code intelligence studies and to encourage its broad use in the community of code intelligence.

\newcolumntype{Y}{>{\raggedright\arraybackslash}X}


\begin{table}[tb]
\centering
\small
\caption{The comparison of released resources between our OpenCoder with other popular open-sourced code LLMs. \textbf{HumanEval scores are reported for the corresponding chat models}.}
\label{table:coder_comparison}
\begin{tabularx}{\textwidth}{p{2.5cm}Xp{1.6cm}Xp{1.5cm} >{\raggedleft\arraybackslash}X >{\raggedleft\arraybackslash}X}
  \toprule
  \textbf{Models} & \textbf{Data}\newline \textbf{Processing Pipeline} & \textbf{Reproducible Pretraining Dataset} & \textbf{Large-scale SFT Dataset ($>$1M)} & \textbf{Intermediate Checkpoints} & \textbf{Training Tokens} & \textbf{HumanEval}\newline \texttt{Pass@1} \\
  \midrule
  \multicolumn{7}{c}{\textit{Open Model Weights \& Reproducible Datasets}}\\ 
  {OpenCoder-8B} & \cmark & \cmark & \cmark & \cmark & 2.5T & 83.5 \\
    StarCoder2-15B & \cmark & \cmark & \xmark & \xmark & 4.1T & 72.6 \\
  Crystal-7B & \xmark & \cmark & \xmark & \cmark & 1.3T & 34.1 \\
  \midrule
    \multicolumn{7}{c}{\textit{Open Model Weights}}\\ 
  CodeLlama-7B & \xmark & \xmark & \xmark & \xmark & 2.5T & 34.8 \\
  CodeGemma-7B & \xmark & \xmark & \xmark & \xmark & 6.5T & 56.1 \\
  DS-Coder-V2-Lite & \xmark & \xmark & \xmark & \xmark & 10.2T & 81.1 \\
  Yi-Coder-9B & \xmark & \xmark & \xmark & \xmark & 6.0T & 85.4 \\
  Qwen2.5-Coder-7B & \xmark & \xmark & \xmark & \xmark & 23.5T & 88.4 \\
  \bottomrule
\end{tabularx}
\end{table}  

\section{Pretraining Data}

Pretraining data plays a crucial role in the development of LLMs, where the scale, quality, and diversity of the data greatly affect the model's overall performance. 
Therefore, we introduce an efficient and effective methodology for producing data tailored for our code LLM pretraining. In this section, we will comprehensively illustrate the data processing strategies used in both the general pretraining stage and the annealing stage.

\subsection{RefineCode}
Pretraining data forms the foundation for the capabilities of large language models. In the LLM open-source community, The Stack v2~\citep{lozhkov2024starcoder} has provided a valuable code dataset, which significantly facilitates the training of code LLMs. However, the quality of the training part in The Stack v2 is insufficient to train LLMs with top-rated performance. 
To address this, we present \textbf{RefineCode}, a high-quality,  reproducible dataset of 960 billion tokens across 607 programming languages, incorporating over 130 language-specific rules with customized weight assignments. 
This dataset is composed of two main parts: raw code and code-related web data. Specifically, we collect the raw code primarily from GitHub repositories up to November 2023 with non-GitHub data from The Stack v2. Additionally, the code-related web data is primarily sourced from web corpora. A detailed comparison with previous versions of The Stack is provided in the Appendix~\ref{appendix:comparison_of_datasets}.
Besides, to ensure both quality and diversity, as shown in Figure~\ref{fig:code_data_pipeline}, we have designed a sophisticated data processing pipeline to produce code pretraining corpus. In the following sections, we have provided a detailed description of our processing pipeline and the details of our \textbf{RefineCode} dataset.

\subsubsection{Raw Code}
To ensure the curation of high-quality raw code data, we have developed the code-specific data processing pipeline including modules of \textbf{preprocessing}, \textbf{deduplication}, \textbf{transformation}, \textbf{filtering}, \textbf{data sampling}. The following sections provide the details of these processes.

\paragraph{Preprocessing}
Initially, we exclude files exceeding 8 MB in size, as these are predominantly non-text files, which require considerable resource overhead. Furthermore, given the miscellaneous file types present on GitHub, we restrict our selection to those file types related to programming languages by their file extension referring to \textit{linguist}\footnote{\small \url{https://github.com/github-linguist/linguist/blob/main/lib/linguist/languages.yml}}, and filter those types with low capacity or low quality. Finally, we preserve 607 different types of programming language files. A comprehensive list of the included and excluded programming languages is provided in Appendix~\ref{appendix:program_langs}.  

\paragraph{Deduplication}
The purpose of deduplication is to construct an unbiased and diverse training set while significantly reducing the data volume. 
Owing to the extremely high repetition of the source code in Github, we prioritize the deduplication process early in the pipeline and adopt an aggressive file-level deduplication strategy (see elaborate analysis in Section \ref{section:code_dedup_analysis}). More specifically, we leverage both exact deduplication and fuzzy deduplication methods to eliminate documents containing identical or near-identical code content shown as follows:

\begin{figure*}[t]
    \centering
    \includegraphics[width=1.0\textwidth]{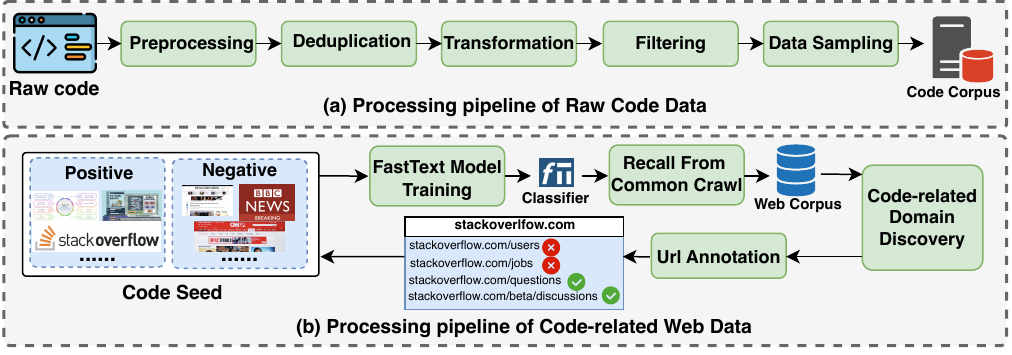} %
    \caption{The illustration of our pretraining data processing workflow.}
    \label{fig:code_data_pipeline}
\end{figure*}

\textit{Exact Deduplication}: Due to the prevalence of forking and copy-pasting within the codebase, nearly 75\% of files are completely duplicated. On account of this, differing from general deduplication process, Identity removal is applied towards code data at the first step in this module. We compute the SHA256 hash value for each document,
where files with identical hash values are compared, and only the code files with the highest star count as well as the latest commit time are retained. 

\textit{Fuzzy Deduplication}: Following the fuzzy deduplication setting in the general data pipeline, we split the raw text into 5-gram pieces, and then calculate the 2048 MinHash functions~\citep{DBLP:conf/sequences/Broder97}. Additionally, we utilize LSH~\citep{DBLP:books/cu/LeskovecRU14} by setting bands to 16 and rows to 128, to retain only those distinct files with the highest stars and latest commit time. This process removes 6\% file volume.

\paragraph{Transformation}
Filtering is generally adequate for removing files that fail to meet specific criteria. However, certain issues, though small in text size, are pervasive across numerous files. In such cases, it is unacceptable to exclude all those issued files. Instead, we opt to transform these files to rectify the identified issues before the filtering module. Concretely, we implement two types of transformation rules as follows:

\textit{Copyright Removal}:
There are over 15\% code files including the copyright notices at the beginning of the content like ``Copyright Intel Corporation (C) 2014-2016'', which are highly repetitive and irrelevant to the coding tasks, possibly affecting the performance of the LLM. Consequently, we specifically identified and removed these copyright notices from the initial code comments.

\textit{PII Reduction}: 
Personally Identifiable Information (PII) encompasses content such as passwords, emails, IP addresses. Training on those data containing PII implies significant privacy risks. 
Therefore, we employ complex regular expressions to detect such information and replace them with placeholders such as ``<name>'' and ``<password>''.

\paragraph{Filtering}
The quality of the original code files on GitHub exhibits significant variability, where lower-quality code potentially hinders the LLM pretraining process. Given the distinct nature of code compared to natural language, the criteria for high-quality code differ significantly from those for natural language. Furthermore, different programming languages also exhibit distinct properties. Based on this, we believe that designing a set of detailed heuristic filtering rules tailored specifically to the characteristics of pretraining data is important to enhance the model's capabilities. Drawing inspiration from the principles of high-quality code data proposed in~\citet{gunasekar2023textbooks}, we consider the following guidelines when designing our filters: \textbf{1) Filter out files with poor self-containment; 2) Filter out files with poor or minimal logical structure; 3) Remove files that deviate significantly from standard formatting.} 


Based on these guidelines and the characteristics of our dataset, our work presents the first heuristic filtering framework by considering the unique characteristics of different programming languages. Based on RedPajama~\citep{together2023redpajama}, this framework extends and refines the existing rules from StarCoder~\citep{li2023starcoder} to better align with the unique properties of code datasets, resulting in more precise and higher-quality data cleansing. We developed the following three categories of filtering rules:  

\begin{enumerate}
    \item \textbf{Natural Language Filtering Rules}: These rules filter data based on common properties  for all text files, such as file size, number of lines, and other general metrics. Both text and code files share these filtering rules.
    \item \textbf{General Code Filtering Rules}: These rules apply to all code files by filtering data based on general code characteristics, such as the number of variables, average function length, and other common features.
    \item \textbf{Language-Specific Filtering Rules}: These rules are designed according to the unique characteristics of specific programming languages, such as the frequency of ``pass'' statements in Python or the use of ``goto'' statements in C. We have developed these rules for the following eight commonly used programming languages: Python, C, C++, C\#, Java, JavaScript, Go, and HTML.  
\end{enumerate}

Heuristic rules involve extensive threshold setting. When defining these rules and determining thresholds, we consistently follow a guiding principle: to remove harmful data as much as possible, while ensuring the overall distribution of the dataset is not significantly affected. We outline our motivations for rule design in Appendix~\ref{appendix:filter_design}, along with a detailed explanation of the tuning process for the corresponding thresholds. Besides, we show the details of several representative rules in Appendix~\ref{appendix:filter_rules}.

\paragraph{Data Sampling}
We try to preserve the original data distribution as much as possible to maximize the utilization of our cleaned high-quality dataset.
However, we downsample certain high-resource programming languages before using our dataset in pretraining. Specifically, we downsample Java data from 449GB to 200GB, due to its excessive volume compared to other common languages. Additionally, we downsample HTML data from 474GB to 64GB, as HTML files often contain a significant amount of non-informative structured content and lack substantial coding logic. Finally, we produce about 730B tokens in the pretraining stage. 

\begin{figure}[tb]
\centering
\includegraphics[width=\textwidth]{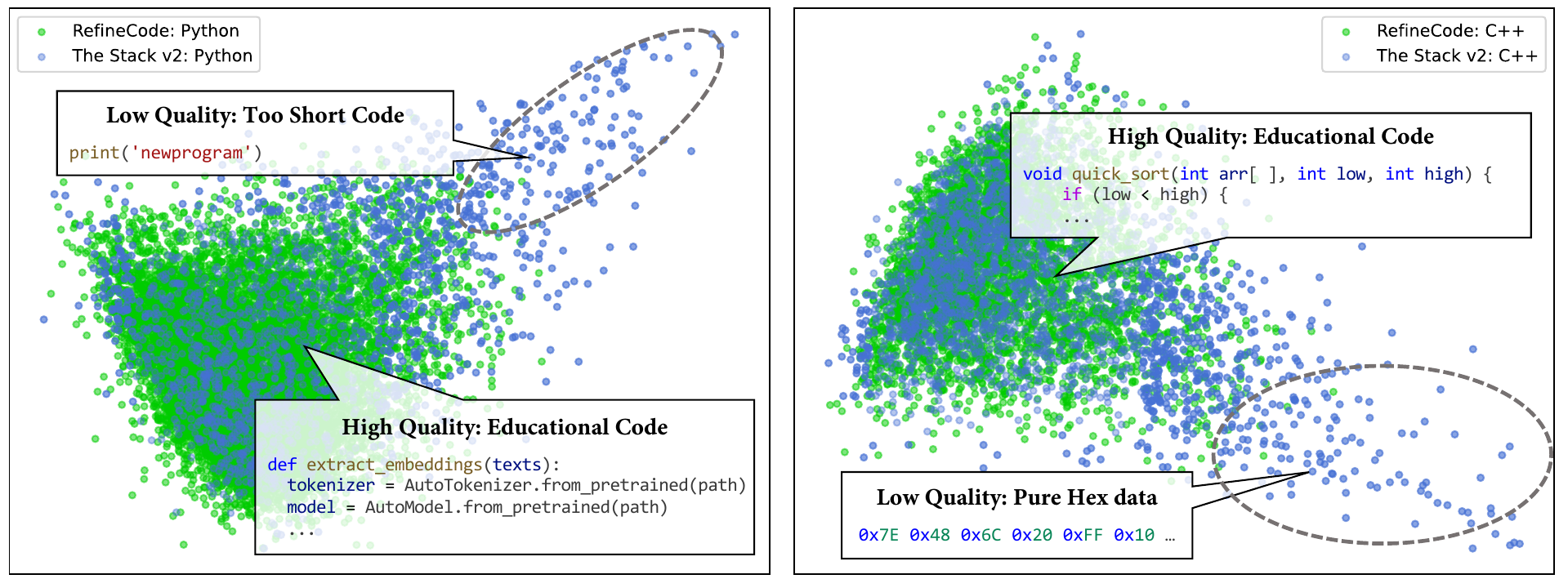}
\caption{Visualization on the PCA data distributions of \textbf{RefineCode} and The Stack v2.}
\label{fig:vis_embedding}
\end{figure}

Notably, as illustrated in Figure~\ref{fig:vis_embedding},
we use PCA to visualize the embeddings extracted from CodeBERT~\citep{feng2020codebert} for The Stack V2 and RefineCode, and observe a clear distinction between these datasets.
Specifically, in Figure~\ref{fig:vis_embedding}, The Stack V2 data shows a greater number of outliers, while the embeddings of RefineCode appear more tightly clustered. 
Besides, after analyzing the outlier data, we observe the outliers usually show many low-quality patterns, such as pure text comments, hexadecimal-only data, and excessively short code lacking computational logic,
which can distort the distribution of the pretraining dataset and ultimately hinder the efficiency of pretraining.

\subsubsection{Code-Related Web Data}

Inspired by the DeepSeekMath~\citep{deepseek-math}, we collect high-quality code-related data corpus from the Common Crawl dataset. Unlike the previous practice in the math domain, due to the lack of open-source fine-gained code corpus, we first annotate 500,000 high-quality code-like data from CommonCrawl using the Autonomous Data Selection~\citep{automathtext} method as seed data for training FastText\citep{fastetxt}. These data serve as the initial code seed corpus.

As shown in Figure~\ref{fig:code_data_pipeline}, the processing pipeline of code-related web data comprises four main components: 1) \textbf{FastText Model Training:} To maintain a controllable vocabulary size in fastText and enable tokenization of Chinese texts using spaces, we first apply the BPE (Byte Pair Encoding) tokenizer to segment the corpus. Subsequently, the open-source FastText framework is utilized for model training. 2) \textbf{Recall From Common Crawl: } We perform recall on Common Crawl to generate the code-related web corpus.
3) \textbf{Code-related Domain Discovery}, we conduct statistical analysis of the recalled data by domain URLs, and define a domain as web pages with the same base URL(e.g. stackoverflow.com),
where domains with over 10\% of web pages are classified as code-related. 
Note that given the scarcity of Chinese data, we provide detailed annotations of domain names related to code and mathematics within the CommonCrawl dataset in the Appendix~\ref{appendix:extra_data}.
4) \textbf{Url Annotation:} We manually annotate the URLs associated with code content within these identified domains. For instance, we have identified all content under ``stackoverflow.com/questions''  as computer technology questions. Then, we include samples with URLs matching ``stackoverflow.com/questions'', 
which are not correctly classified by fastText, into our code seed corpus. After three iterations, we obtain about 220GB code-related web data. 
Note that as the iteration progresses, the quantity and diversity of the seed corpus will be better.

We also apply the same recall pipeline to FineWeb~\citep{penedo2024fineweb}, Skypile~\citep{wei2023skywork} and web part of AutoMathText~\citep{automathtext} and produce 330GB code-related web data in total. 
Furthermore, we observe that only a very small portion of the textual data in GitHub is also related to natural language text.
Therefore, we also train a classifier to determine whether the text is code-related and obtain an additional 178GB code-related web data.


\subsubsection{Summary}

\begin{figure*}[h!]
    \centering
    \includegraphics[width=\textwidth]{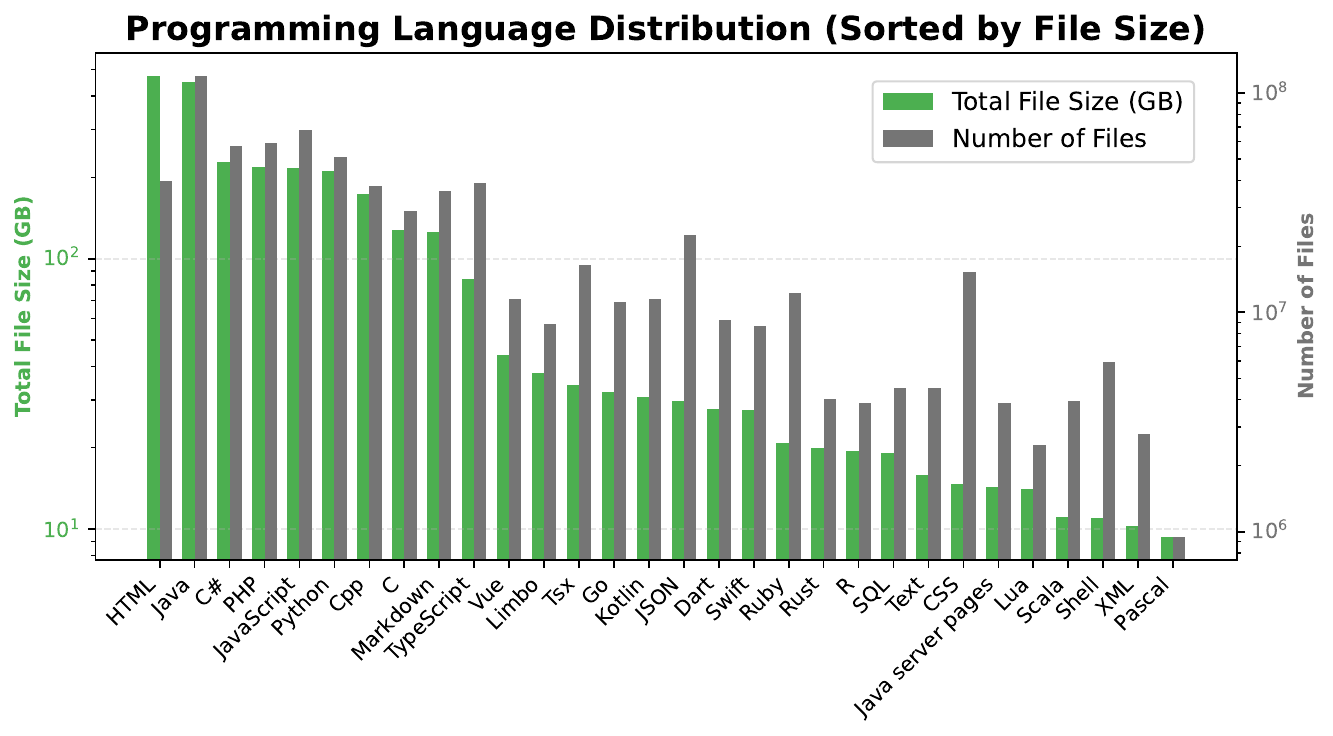} %
    \caption{The distribution of top program languages in \textbf{RefineCode} (before data sampling).}
    \label{fig:code_data_dis}
\end{figure*}

Ultimately, we curated a high-quality code pretraining dataset, \textbf{RefineCode}, consisting of about 960 billion tokens. The composition of the data sources is illustrated in Table~\ref{table:corpus_composition}, while the distribution of different program languages is displayed in Figure~\ref{fig:code_data_dis}. For more details regarding the data composition of different program languages, please refer to Appendix~\ref{appendix:data_composition}. To demonstrate the efficacy of RefineCode, we train a 1.5B code LLM up to 600B using data from RefineCode and the training subset of The Stack v2 respectively. The results in Figure~\ref{fig:enter-label}, indicate that RefineCode significantly improves training efficiency compared to The Stack v2, highlighting the superiority of our dataset.

\begin{table}[tb]
\centering
\caption{The Composition of \textbf{RefineCode}.}
\label{table:corpus_composition}
\resizebox{0.8\textwidth}{!}{
\begin{tabular}{llrr}
  \toprule
  \textbf{Category} & \textbf{Data Source} & \textbf{\# Tokens} & \textbf{Percentage} \\
  \midrule
  \multirow{3}{*}{Raw Code Data} & Github Code & 755 B & 78.4\% \\
        & Jupyter Notebooks & 11 B & 1.1\% \\
        & The Stack v2 & 120 B & 12.5\% \\
  \midrule
  \multirow{3}{*}{Code-related Web Data} & Processed CC & 13 B & 1.4\% \\
       & Processed SkyPile & 3 B & 0.3\% \\
       & Processed FineWeb & 55 B & 5.7\% \\
  \midrule
  OpenSource Data & Processed AutoMathText & 3 B & 0.3\% \\
  \bottomrule
\end{tabular}
}
\end{table}

\subsection{Annealing Data}
The annealing stage can be seen as a bridge between the general pretraining stage and the supervised fine-tuning (SFT) stage. Following the training strategy in MiniCPM~\citep{hu2024minicpm}, our model also undergoes a rapid learning rate annealing phase after the general pretraining stage,
where very high-quality training data is used to further enhance the model's capabilities. In addition to the RefineCode from the original distribution, we further incorporated the Algorithmic Corpus and synthetic data during the annealing phase. The detailed data mixture can be found in Table \ref{table:decay_data_mixture}.

\paragraph{Original Distribution Data} In the annealing stage, it's necessary to ensure that the overall data distribution remains similar to the pretraining phase. A significant distribution shift can lead to catastrophic forgetting in the model's knowledge,
and we ensure that 84\% of the annealing data comes from the original distribution of RefineCode. Note that given the limited computing budget available, this mixture ratio might not be ideal.

\paragraph{Algorithmic Corpus}
Algorithmic code files exhibit strong code logic and minimal dependency on external files, demonstrating excellent self-containment. Additionally, these files are more aligned with the distribution of smaller, independent tasks commonly encountered in real-world interactive scenarios. 
Therefore, we sample a certain proportion of the original pretraining data that contains keywords such as ``leetcode,'', ``def solution,'' or ``class solution'' to create this corpus.

\paragraph{Synthetic Data}
High-quality pretraining data rewriting is also extremely important during the pretraining stage, which helps the model memorize and embed knowledge for efficient retrieval~\citep{allen2023physics}. We select Algorithmic Corpus as the seed because it encompasses a wide range of algorithmic logic. We employed two forms of rewriting enhancement: Verified Code Snippets and Code Textbooks. 

\begin{enumerate}
    \item \textbf{High Quality Code Snippet:} Inspired by the synthetic CodeExercises dataset in~\citet{gunasekar2023textbooks}, we utilized the algorithmic corpus as seeds and employ a strong LLM to synthesize a batch of self-contained independent functions along with their corresponding test cases. We retained the data that successfully passed the test cases and included them in the annealing stage dataset. This approach was similarly extended to support multiple program languages.

    \item \textbf{Code Textbooks: } To enable the model to understand code from multiple perspectives, we constructed educational text snippets based on the hqcode~\footnote{\url{https://huggingface.co/datasets/yuxiang630/hqcode}} dataset using Qwen2-72B-Instruct~\citep{yang2024qwen2}. Hqcode is a multilingual code dataset synthesized with GPT-4o-Mini, where each entry describes an independent task and provides a corresponding function as a solution. We engaged LLMs to perform interactive analysis on the code within this dataset, extracting and elaborating on abstract code knowledge. This approach aims to enable the model to learn code from diverse perspectives.
    
\end{enumerate}



\begin{table}[h]
\centering
\caption{Detailed data mixture for annealing data. }
\label{table:decay_data_mixture}
\begin{tabular}{llr}
  \toprule
  \textbf{Category} & \textbf{Dataset} & \textbf{\# Token} \\
  \midrule
    \multirow{2}{*}{Original Data} & RefineCode & 83.94 B\\
    & Algorithmic Corpus & 12.44 B \\
    \midrule
     \multirow{2}{*}{Synthetic Data} & High Quality Code Snippet & 2.71 B  \\
    & Code Textbooks & 0.91 B  \\
  \bottomrule
\end{tabular}
\end{table}

\section{Pretraining}

\subsection{Model Architecture}

In this section, we provide a detailed overview of our model architecture. As shown in Table \ref{model_archi_table}, the models are available in two sizes: 1.5 billion and 8 billion parameters. The 1.5 billion model consists of 24 layers with a hidden size of 2240, 14 attention heads, and 14 key/value heads, supporting a context window size of 4096. The 8 billion model architecture closely follows the Llama-3.1-8B architecture, with 32 layers, a hidden size of 4096, and 8 attention heads. Both models use the SwiGLU activation function and have a vocabulary size of 96,640, using the tokenizer proposed in~\citet{inf-llm}.


\begin{table}[t]
    \centering
    \caption{Overview of the key hyperparameters of OpenCoder, including 1.5B and 8B. }
    \label{model_archi_table}
    \begin{tabular}{lcc}
        \toprule
        & \textbf{OpenCoder-1.5B} & \textbf{OpenCoder-8B} \\
        \midrule
        {Layers} & 24 & 32 \\
        {Model Dimension} & 2240 & 4096 \\
        {Attention Heads} & 14 & 32 \\
        {Key / Value Heads} & 14 & 8 \\
        {Activation Function} & \multicolumn{2}{c}{SwiGLU} \\
        {Vocab Size} & \multicolumn{2}{c}{96640} \\
        {Positional Embedding} & RoPE($\theta = 10000$) & RoPE($\theta = 500000$) \\
        {Context Window Size} & 4096 & 8192  \\
        \bottomrule
    \end{tabular}
\end{table}

\subsection{Training Details}
The training process, based on the aforementioned model architecture, involved several critical details. The dataset encompassed both Chinese and English languages, alongside 607 programming languages, the complete list of which is provided in Appendix \ref{appendix:program_langs}. 

For the 1.5B model, due to the incomplete data curation, training was performed on 2 trillion tokens over four epochs. Following the pretraining phase, we conducted annealing training on an additional 100 billion tokens. The WSD learning schedule, referenced in MiniCPM~\citep{hu2024minicpm}, was employed, featuring a warm-up phase of 2,000 steps across 8 billion tokens. The peak learning rate was 3e-4, which remained constant after the warm-up and subsequently decayed exponentially to 1e-5 during the annealing phase. A micro-batch size of 4 and a global batch size of 1024 were used. Training was conducted using Megatron-LM~\citep{shoeybi2020megatronlmtrainingmultibillionparameter} with distributed optimization and DDP gradient overlap on a cluster of 256 H800 GPUs over a total of 109.5 hours, equating to 28,034 GPU hours.

For the 8B model, the WSD learning schedule was again employed with a warm-up phase covering 8 billion tokens over 2,000 steps. This model was trained for 3.5 epochs on 2.5 trillion tokens, followed by a decay phase with an additional 100 billion tokens. Unlike the 1.5 billion model, which lacked code-related recall data due to incomplete data processing, the 8 billion model incorporated this data during training.
The learning rate schedule mirrored that of the 1.5B model. The micro-batch size was set to 1, with a TP of 2 and a sequence length of 8192. The global batch size was 1024. Training was conducted on a cluster of 512 H100 GPUs over 187.5 hours, totaling 96,000 GPU hours. It is noteworthy that the first 130,000 steps were trained with a sequence length of 4096 and a global batch size of 2048.

\section{Post Training}
\subsection{Data Composition}

\paragraph{Open-source Training Data}
To enhance the model training, we collect the open-source instruction corpora from the websites, including Evol-Instruct\footnote{\small \url{https://huggingface.co/datasets/theblackcat102/evol-codealpaca-v1}}~\citep{wizardcoder}, Infinity-Instruct\footnote{\small \url{https://huggingface.co/datasets/BAAI/Infinity-Instruct}}, McEval\footnote{\small \url{https://huggingface.co/datasets/Multilingual-Multimodal-NLP/McEval-Instruct}}~\citep{mceval,wmt2021}, where the instruction data is created from the multilingual raw code snippet by language sampling with the fixed ratio.
We employ an LLM to perform binary classification on the content of Infinity-Instruct, aiming to extract the segments specifically related to the code. Additionally, we sample real user queries from WildChat~\citep{zhao2024wildchat} and Code-290k-ShareGPT\footnote{\small \url{https://huggingface.co/datasets/cognitivecomputations/Code-290k-ShareGPT-Vicuna}}, extracting code-related dialogue histories using LLM and subsequently performing data cleaning. For low-quality responses, we employ a robust LLM to regenerate the content, enhancing the overall data quality. This RealUser-Instruct dataset not only exhibits high diversity but also aligns more closely with real-world problem complexity, focusing on addressing practical issues in authentic scenarios.

\begin{figure*}[t]
    \centering
    \includegraphics[width=1.0\textwidth]{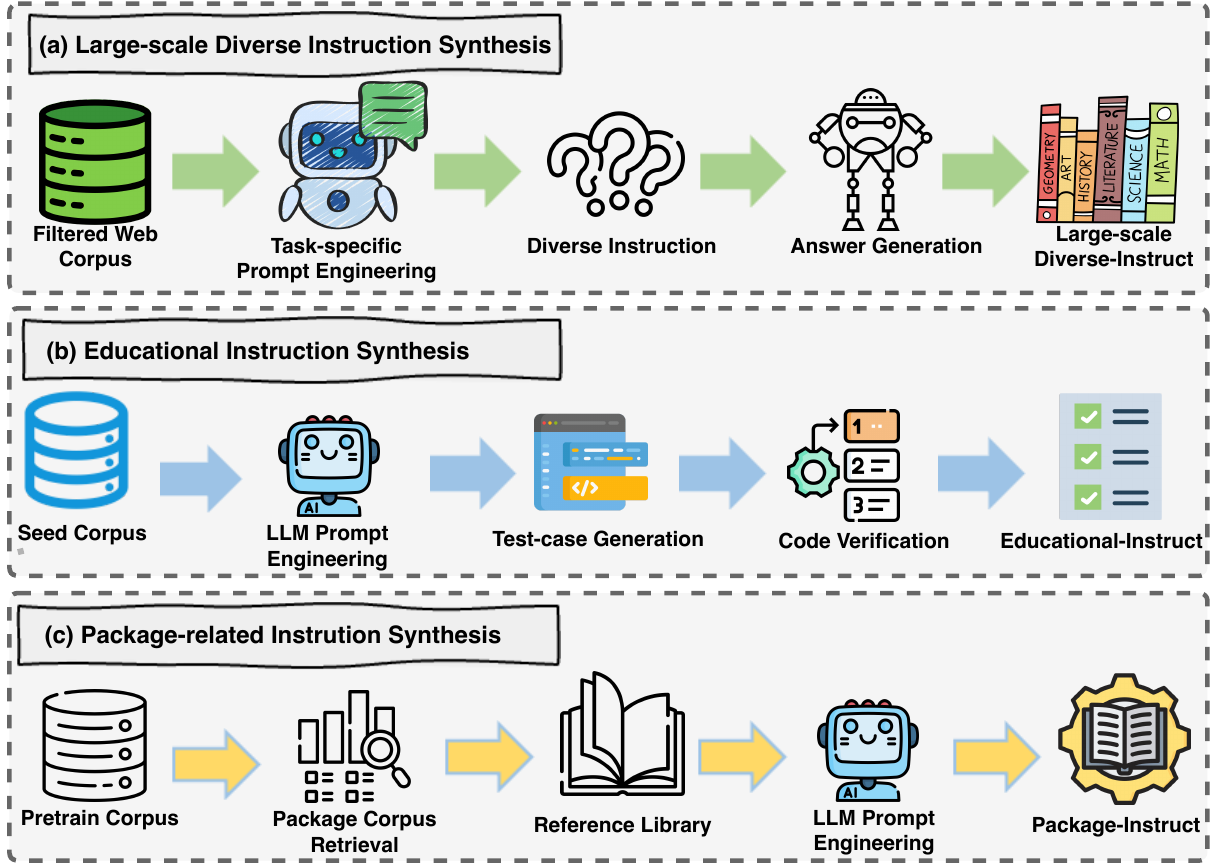} %
    \caption{The illustration of our instruction data synthesis workflow.}
    \label{fig:sft_data_pipeline}
\end{figure*}

\paragraph{Educational Instruction Synthesis}
To ensure the diversity and richness of instruction-tuning datasets, prior work explores using code snippets sampled from real-world sources as seed data~\citep{wei2023magicoder}, subsequently used to synthesize question-answer pairs. This approach is widely adopted in the development of large language models. In synthesizing instruction-tuning datasets for Python code, we enhance the effectiveness of this method. Specifically, we observe that the educational value of the synthesized data largely depends on the quality of the seed data. Thus, during the seed data selection phase, we use a scorer model where the input is a code snippet to identify high-quality seed data. By using only high-quality seed data, we ensure that the resulting instruction-tuning dataset includes more educational example responses. Subsequently, we use a teacher model to generate multiple test cases for the code sections in each problem. These test cases are appended to the code snippets and executed using a Python interpreter. Only the data samples that successfully pass the tests are retained. By using this strategy, we maximize the likelihood that the generated data is both syntactically and semantically sound, thereby enhancing the reliability of the dataset.

\paragraph{Package-related Instruction Synthesis}


Due to a significant amount of outdated package usage in the pre-training data, LLM may sometimes employ methods from older versions of libraries when generating code, leading to suboptimal performance in tasks involving package invocation. 
For example, Python's extensive ecosystem of libraries—such as NumPy, pandas, and TensorFlow—are frequently updated, with new functions, methods, and best practices emerging over time. As a result, when users query NumPy, the model may give incorrect answers based on outdated information. Furthermore, if the model is significantly affected by outdated library syntax, it may fail to generate correct code, leading to errors when the code is executed in a Python interpreter.
This problem undermines the model's ability to use tool calls to improve performance. To mitigate the impact of outdated programming syntax and obsolete external library interfaces in the pre-training dataset, we synthesized a tool usage instruction tuning dataset using up-to-date external library documentation. Specifically, we analyzed commonly used external Python libraries and retrieved API signatures and usage examples for widely used syntax and tools via PyDoc. This information was sent to prompt a teacher model that generated accurate and up-to-date question-answer pairs reflecting current usage. 
By fine-tuning the model on a curated set of code that includes up-to-date usage of these libraries, we ensured that it could provide accurate, contemporary answers to questions about using them effectively. This is particularly important given the rapid pace of change in software development, where outdated code and obsolete practices can lead to incorrect answers and inefficient solutions.

\paragraph{Large-scale Diverse Instruction Synthesis}
Following the previous work~\citep{yue2024mammoth2}, to increase the diversity of the instruction dataset, we create a large-scale instruction data synthesis framework.
The framework for synthesizing code instruction data using LLMs incorporates the following key components: 
(1) An LLM is used first to clean the irrelevant context (e.g. advertisements on the web) in the websites and select useful sentences as the seed for further question generation.
(2) A task specification module defines programming languages, difficulty levels, and coding task types, utilizing a configuration file for easy customization. The prompt engineering component employs a template-based system to generate diverse, contextually rich prompts, incorporating real-world scenarios and best practices in software development. We set temperature $T=1.0$ for diverse questions.
(3) An advanced LLM with more parameters first generates the created questions and then generates the corresponding answers. The validation module combines automated code execution and unit testing to check the correctness. 
(4) Then an LLM is adopted to refine the response by adding code comments and more explanation. 

\subsection{Two-Stage Instruction-Tuning}

In developing a code LLM, particularly in computer science and software development, it is essential to ensure that the model excels in both theoretical knowledge and practical coding tasks. To address both needs, we implemented a two-stage instruction fine-tuning process. The detailed composition of instruction tuning is presented in Table \ref{table:sft_composition}.

The first stage of this fine-tuning process focused on synthesizing question-answer (QA) pairs related to theoretical computer science. Building on general-purpose pre-training data, we created a specialized dataset that enabled the model to develop a deeper understanding of theoretical computer science, such as algorithms, data structures, and networking principles. By fine-tuning the model with domain-specific QA pairs, we ensured that it could respond with greater precision to questions about concepts such as binary search trees, dynamic programming, and the intricacies of object-oriented design patterns.

In the second stage of the fine-tuning process, we shifted focus from theoretical knowledge to practical coding tasks. In this stage, we used high-quality code from GitHub to create a dataset aimed at improving the model’s ability to generate and work with code. By fine-tuning the model on high-quality code from GitHub, we ensured it was exposed to real-world examples of well-maintained and formatted code. One key advantage of using high-quality code in the fine-tuning process is that it enhances the model’s ability to generate code that is both syntactically and semantically correct.

The two-stage fine-tuning approach allows the model to excel in theoretical knowledge and practical coding tasks, thereby avoiding the limitations of focusing on only one area. Models that only prioritize theory may struggle with coding, while those focused solely on code generation may lack depth in explaining complex concepts. By refining both areas, the model becomes technically proficient and versatile, able to meet the needs of developers, beginners, and professionals alike.

\begin{table}[tb]
\centering
\caption{Detailed data composition of our two-stage instruction-tuning.}
\label{table:sft_composition}
\begin{tabular}{llr}
  \toprule
  \textbf{Stage} & \textbf{Data Source} & \textbf{\# Examples} \\
  \midrule
  \multirow{3}{*}{Stage1} & RealUser-Instruct & 0.7 M \\
   & Large-scale Diverse-Instruct & 2.3 M \\
   & Filtered Infinity-Instruct & 1.0 M \\
   \midrule
   \multirow{4}{*}{Stage2} & McEval-Instruct & 36 K \\
    & Evol-Instruct & 111 K \\
    & Educational-Instruct & 110 K \\
    & Package-Instruct & 110 K \\
  \bottomrule
\end{tabular}
\end{table}

\subsection{Training Details}
In the first stage of SFT, we trained for one epoch with a batch size of 4096, a learning rate (LR) of 2e-5, warmup steps set to 100, and a cosine learning rate scheduler.In the second stage of SFT, we trained for three epochs using a batch size of 512, a learning rate of 5e-5, with 100 warmup steps, and the same cosine learning rate scheduler.

\subsection{Decontamination}
We applied strict data deduplication for all SFT data. Specifically, we removed any data containing the entry points corresponding to test sets such as HumanEval~\citep{chen2021evaluating} and MBPP~\citep{austin2021program}. Additionally, we performed 10-gram deduplication, removing any data with a 10-gram overlap with the test sets.
\section{Experimental Results}

In this section, we conduct a comprehensive and fair evaluation to demonstrate that the model we constructed using cleaned and synthesized data performs comparably to other closed large language models. We also compared the most widely used and powerful open-source language models, including the Crystal and StarCoder series. To further highlight the practicality and effectiveness of our models, we focus on tasks such as code generation, code completion, and code understanding.

\subsection{Evaluation on Base Models}
For base models, we focus on evaluating their code completion ability. Code completion is a fundamental capability that enables code models to tackle complex tasks. This evaluation goal aligns with our optimization objective in the annealing stage, as code completion can be regarded as a special case of the code generation task. To ensure the reproducibility of all results, we used publicly available LLM evaluation framework \texttt{OpenCodeEval}\footnote{\url{https://github.com/richardodliu/OpenCodeEval}}. For comparing models, we compare open-coder-1.5B with state-of-the-art small language models.


\paragraph{HumanEval \& MBPP} We selected two widely used code completion benchmarks to evaluate OpenCoder, HumanEval~\citep{chen2021evaluating}, and MBPP~\citep{austin2021program}. To further enhance the accuracy of the evaluation, EvalPlus~\citep{liu2024your} extends HumanEval and MBPP into HumanEval+ and MBPP+ by adding unique and challenging test cases and correcting inaccurate ground-truth solutions. These results can be used to indicate the model’s ability to understand and apply basic Python data structures and knowledge of algorithms. For HumanEval, we report the 0-shot results. For MBPP, we report 3-shots on 500 questions in the test split from original dataset, while the others following EValPlus report results on 378 questions in the sanitized part.

\paragraph{BigCodeBench} 

BigCodeBench~\citep{zhuo2024bigcodebench} is a challenging benchmark for code completion, designed to assess models on their ability to handle complex instructions and make accurate function calls across diverse external libraries. In the Completion setup, models are provided with a function signature and related documentation to generate appropriate code, along with a unit test for the completed function. Covering a range of practical programming tasks, it evaluates models’ ability to handle real-world scenarios involving complex, task-specific libraries.

\begin{table*}[t]
\centering
\caption{Performance of various base models on HumanEval, MBPP, and the ``complete'' task of BigCodeBench. Models trained on reproducible datasets are marked with {\color{green!88}{\textbf{green}}}.}
\resizebox{\textwidth}{!}{
\begin{tabular}{lr|cc|ccc|cc}
\toprule
\multirow{2}{*}{\textbf{Model}} & 
\multirow{2}{*}{\textbf{Size}} & 
\multicolumn{2}{c|}{\textbf{HumanEval}} & 
\multicolumn{3}{c|}{\textbf{MBPP}} &
\multicolumn{2}{c}{\textbf{BigCodeBench}} \\
& & 
\textit{HE} & \textit{HE+} & 
\textit{MBPP} & \textit{MBPP+} & \textit{3-shot} &
\textit{Full} & \textit{Hard}  \\
\midrule
\multicolumn{9}{c}{\textbf{1B+ Models}} \\
\midrule
DeepSeek-Coder-1.3B-Base & 1.3B & 34.8 & 26.8 & 55.6 & 46.9 & 46.2 & 26.1 & 3.4 \\
Yi-Coder-1.5B & 1.5B & 41.5 & 32.9 & 27.0 & 22.2 & 51.6 & 23.5 & 3.4 \\
CodeGemma-2B & 2B & 31.1 & 16.5 & 51.1 & 43.1 & 45.4 & 23.9 & 7.4 \\
Qwen2.5-Coder-1.5B & 1.5B & 43.9 & 36.6 & 69.2 & 58.6 & \textbf{59.2} & \textbf{34.6} & \textbf{9.5} \\
\rowcolor{green!8} StarCoder2-3B & 3B & 31.7 & 27.4 & 60.2 & 49.1 & 46.4 & 21.4 & 4.7 \\
\rowcolor{green!8} {OpenCoder-1.5B-Base} & 1.5B & \textbf{54.3} & \textbf{49.4} & \textbf{70.6} & \textbf{58.7} & 51.8 & 24.5 & 5.4 \\
\midrule

\multicolumn{9}{c}{\textbf{6B+ Models}} \\
\midrule
CodeLlama-7B & 7B & 33.5 & 26.2 & 55.3 & 46.8 & 41.4 & 28.7 & 5.4 \\
CodeGemma-7B & 7B & 39.0 & 32.3 & 50.5 & 40.7 & 55.0 & 38.3 & 10.1 \\
DS-Coder-6.7B-Base & 6.7B & 47.6 & 39.6 & 70.2 & 56.6 & 60.6 & 41.1 & 11.5 \\
DS-Coder-V2-Lite-Base (MoE) & 16B & 40.9 & 34.1 & 71.9 & 59.4 & 62.6 & 30.6 & 8.1 \\
CodeQwen1.5-7B & 7B & 51.8 & 45.7 & 72.2 & 60.2 & 61.8 & 45.6 & 15.6 \\
Yi-Coder-9B & 9B & 53.7 & 46.3 & 48.4 & 40.7 & \textbf{69.4} & 42.9 & 14.2 \\
Qwen2.5-Coder-7B & 7B & 61.6 & 53.0 & 76.9 & 62.9 & 68.8 & \textbf{45.8} & \textbf{16.2} \\
\rowcolor{green!8} Crystal-7B & 7B & 22.6 & 20.7 & 38.6 & 31.7 & 31.0 & 10.8 & 4.1\\
\rowcolor{green!8} StarCoder2-7B & 7B & 35.4 & 29.9 & 54.4 & 45.6 & 55.2 & 27.7 & 8.8 \\
\rowcolor{green!8} StarCoder2-15B & 15B & 46.3 & 37.8 & 66.2 & 53.1 & 15.2 & 38.4 & 12.2 \\
\rowcolor{green!8} {OpenCoder-8B-Base} & 8B & \textbf{66.5} & \textbf{63.4} & \textbf{79.9} & \textbf{70.4} & 60.6 & 40.5 & 9.5 \\
\bottomrule
\end{tabular}
}
\end{table*}

\subsection{Evaluation on Instruct Model}

\paragraph{LiveCodeBench} LiveCodeBench is a comprehensive, contamination-free benchmark that assesses the reasoning and problem-solving abilities of highly complex algorithmic tasks. The benchmark is continuously updated with new problems from platforms such as LeetCode, AtCoder, and CodeForces, ensuring the challenges remain current and diverse. LiveCodeBench provides a robust measure of a model's ability to handle sophisticated logical processes, which are essential in competitive programming contexts. The instruct models are evaluated on the 2305-2409 data split.

\begin{table*}[!h]
\centering
\caption{Performance of various chat models on HumanEval, MBPP, the ``instruct'' task of BigCodeBench and LiveCodeBench. Models trained on reproducible datasets are marked with {\color{green!88}{\textbf{green}}}.}
\resizebox{\textwidth}{!}{
\begin{tabular}{lr|cc|cc|cc|c}
\toprule
\multirow{2}{*}{\textbf{Model}} & 
\multirow{2}{*}{\textbf{Size}} & 
\multicolumn{2}{c|}{\textbf{HumanEval}} & 
\multicolumn{2}{c|}{\textbf{MBPP}} &
\multicolumn{2}{c|}{\textbf{BigCodeBench}} &
\multicolumn{1}{c}{\textbf{LiveCodeBench}}
\\
& & 
\textit{HE} & \textit{HE+} & 
\textit{MBPP} & \textit{MBPP+} &
\textit{Full} & \textit{Hard}  & \textit{Avg} \\
\midrule
\multicolumn{9}{c}{\textbf{1B+ Models}} \\
\midrule
DS-coder-1.3B-Instruct & 1.3B & 65.2 & 61.6 & 61.6 & 52.6 & 22.8 & 3.4 & 9.3\\
Qwen2.5-Coder-1.5B-Instruct & 1.5B & 70.7 & 66.5 & 69.2 & 59.4 & 32.5 & 6.8 & \textbf{15.7}\\
Yi-Coder-1.5B-Chat & 1.5B & 67.7 & 63.4 & 68.0 & 59.0 & 24.0 & 6.8 & 11.6\\
\rowcolor{green!8} {OpenCoder-1.5B-Instruct} & 1.5B & \textbf{72.5} & \textbf{67.7} & \textbf{72.7} & \textbf{61.9} & \textbf{33.3} & \textbf{11.5} & 12.8 \\

\midrule
\multicolumn{9}{c}{\textbf{6B+ Models}} \\
\midrule
DS-Coder-V2-Lite-Instruct & 16B & 81.1 & 75.0 & 82.3 & 68.8 & 36.8 & 16.2 & 24.3 \\
CodeLlama-7B-Instruct & 7B & 45.7 & 39.6 & 39.9 & 33.6 & 21.9 & 3.4 & 2.8 \\
CodeGemma-7B-It & 7B & 59.8 & 47.0 & 69.8 & 59.0 & 32.3 & 7.4 & 14.7 \\
DS-Coder-6.7B-Instruct & 6.7B & 78.6 & 70.7 & 75.1 & 66.1 & 35.5 & 10.1 & 20.5 \\
Yi-Coder-9B-Chat & 9B & 82.3 & 72.6 & 81.5 & 69.3 & 38.1 & 11.5 & 23.4  \\
CodeQwen1.5-7B-Chat & 7B & 86.0 & 79.3 & 83.3 & 71.4 & 39.6 & \textbf{18.9} & 20.1 \\
Qwen2.5-Coder-7B-Instruct & 7B & \textbf{88.4} & \textbf{84.1} & \textbf{83.5} & \textbf{71.7} & \textbf{41.0} & 18.2 & \textbf{37.6} \\
\rowcolor{green!8} CrystalChat-7B & 7B & 34.1 & 31.7 & 39.1 & 32.7 & 26.7 & 2.3 & 6.1\\
\rowcolor{green!8} StarCoder2-15B-Instruct-v0.1 & 15B & 72.6 & 63.4 & 75.2 & 61.2 & 37.6 & 12.2 & 20.4 \\
\rowcolor{green!8} {OpenCoder-8B-Instruct} & 8B & 83.5 & 78.7 & 79.1 & 69.0 & 40.3 & 16.9 & 23.2 \\
\bottomrule
\end{tabular}
}
\end{table*}

\paragraph{MultiPL-E} MultiPL-E extends the HumanEval benchmark to evaluate the code generation capabilities of large language models across multiple languages. MultiPL-E translates tasks into languages such as C++, Java, PHP, TypeScript, C\#, Bash, and JavaScript, providing a consistent basis for assessing how models apply their programming skills across different syntaxes and paradigms. We follow the evaluation code of {Qwencoder}\footnote{\url{https://github.com/QwenLM/Qwen2.5-Coder}} to systematically measure performance in each language, providing insights into the adaptability and code generation accuracy of LLMs in a multilingual context.


\begin{table}[t]
\centering
\caption{Performance of various chat models on the MultiPL-E benchmark across different programming languages.}
\label{tab:instruct-multiple}
\resizebox{\textwidth}{!}{
\begin{tabular}{lr|cccccccc|c}
\toprule
\textbf{Model} & \textbf{Size} & 
Python & Java & C++ & C\# & TS & JS & PHP & Bash  & \textbf{Average}\\
\midrule
\multicolumn{11}{c}{\textbf{1B+ Models}} \\
\midrule
DS-Coder-1.3B-Instruct & 1.3B & 65.2 & 51.9 & 45.3 & 55.1 & 59.7 & 52.2 & 45.3 & 12.7 & 48.4 \\
Yi-Coder-1.5B-Chat & 1.5B & 67.7 & 51.9 & 49.1 & 57.6 & 57.9 & 59.6 & 52.2 & 19.0 & 51.9 \\
Qwen2.5-Coder-1.5B-Instruct & 1.5B & 71.2 & 55.7 & \textbf{50.9} & \textbf{64.6} & 61.0 & \textbf{62.1} & \textbf{59.0} & 29.1 & 56.7 \\
\rowcolor{green!8} {OpenCoder-1.5B-Instruct} & 1.5B & \textbf{72.5} & \textbf{64.6} & \textbf{50.9} & 61.4 & \textbf{63.5} & \textbf{62.1} & 55.3 & \textbf{29.7} & \textbf{57.5} \\
\midrule
\multicolumn{11}{c}{\textbf{6B+ Models}} \\
\midrule
DS-Coder-6.7B-Instruct & 6.7B & 78.6 & 68.4 & 63.4 & 72.8 & 67.2 & 72.7 & 68.9 & 36.7 & 66.1 \\
DS-Coder-V2-Lite-Instruct & 16B & 81.1 & \textbf{76.6} & \textbf{75.8} & 76.6 & 80.5 & 77.6&  74.5 & 43.0 & 73.2 \\
CodeLlama-7B-Instruct & 7B & 45.7 & 32.2 & 28.6 & 32.9 & 39.0 & 43.5 & 31.7 & 10.1 & 33.0\\
CodeGemma-7B-It & 7B & 59.8 & 48.1 & 46.6 & 51.9 & 54.7 & 54.0 & 46.6 & 10.1 & 46.5 \\
CodeQwen1.5-7B-Chat & 7B & 83.5 & 70.9 & 72.0 & 75.9 & 76.7 & 77.6 & 73.9 & 41.8 & 71.6 \\
Yi-Coder-9B-Chat & 9B & 85.4 & 76.0 & 67.7 & 76.6 & 72.3 & 78.9 & 72.1 & 45.6 & 71.8 \\
Qwen2.5-Coder-7B-Instruct & 7B & \textbf{87.8} & 76.5 & 75.6 & \textbf{80.3} & \textbf{81.8} & \textbf{83.2} & \textbf{78.3} & \textbf{48.7} & \textbf{76.5} \\
\rowcolor{green!8} {OpenCoder-8B-Instruct} & 8B & 83.5 &  72.2 & 61.5 & 75.9 & 78.0 & 79.5 & 73.3 & 44.3 & 71.0 \\
\bottomrule
\end{tabular}
}
\end{table}

\begin{figure*}[t]
    \centering
    \includegraphics[width=1.0\textwidth]{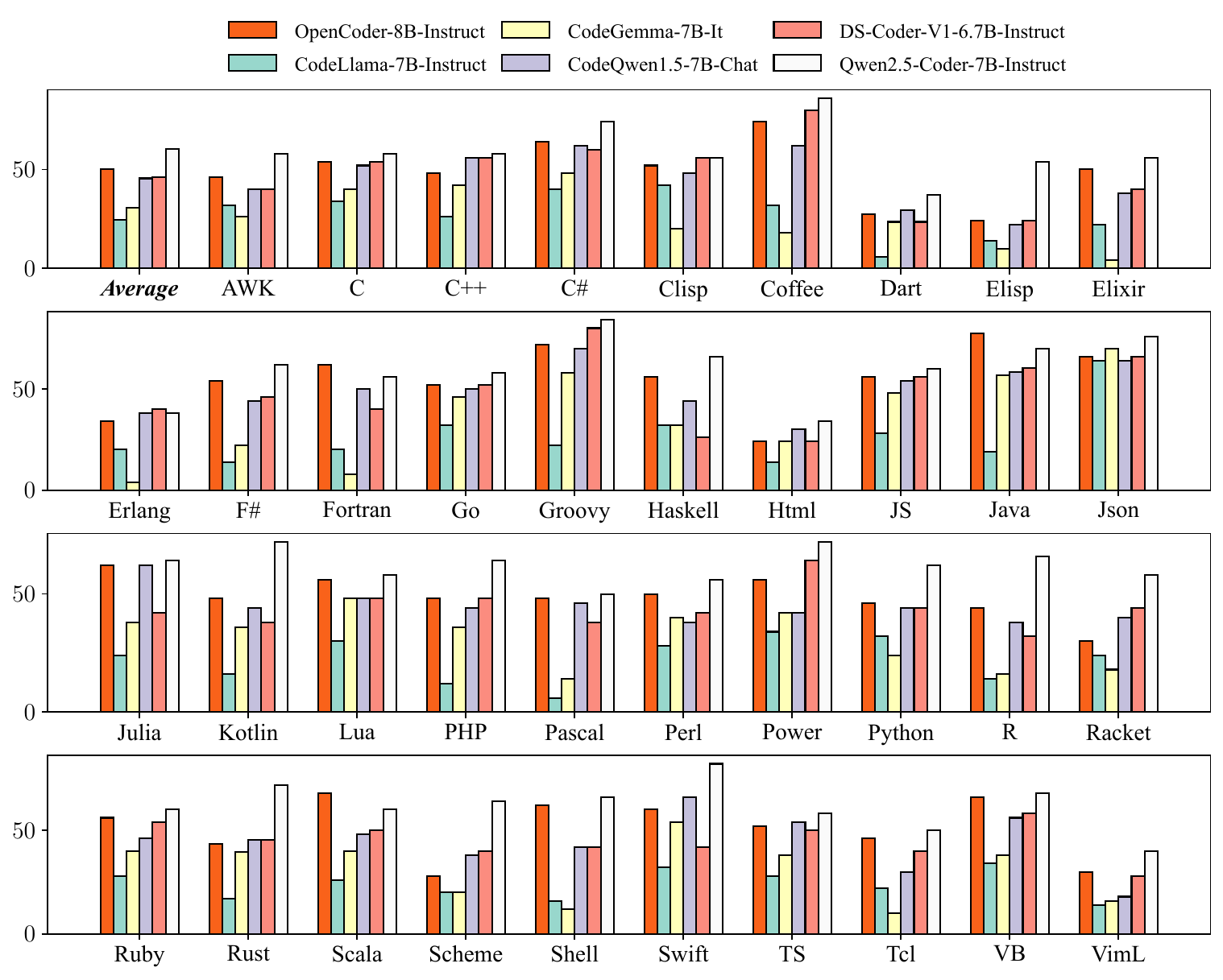} %
    \caption{The McEval performance of OpenCoder-8B-Instruct in comparison to other open-source code models of comparable size.}
    \label{fig:mceval}
    \vspace{-10pt}
\end{figure*}

\paragraph{McEval} The comprehensive multilingual code evaluation benchmark McEval~\citep{mceval} employed a detailed assessment of OpenCoder's programming capabilities across 40 languages. In contrast to MultiPL-E, this benchmark is not derived from HumanEval or MBPP. Figure \ref{fig:mceval} depicts the results of the multilingual generation task for OpenCoder-8B-Instruct, which comprises nearly 2,000 samples. The figure illustrates that the model exhibits superior multilingual performance compared to other open-source models of comparable size.

\begin{figure*}[t]
    \centering
    \includegraphics[width=1.0\textwidth]{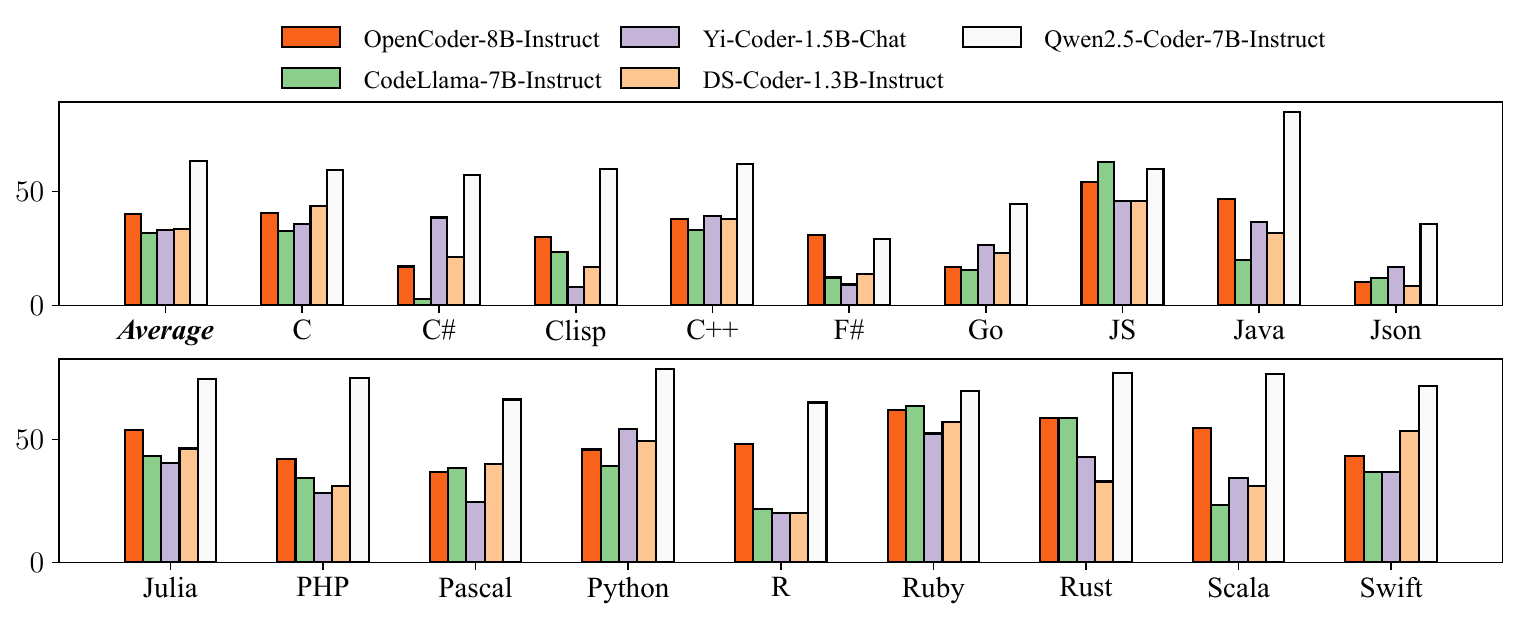} %
    \caption{The MdEval performance of OpenCoder-8B-Instruct in comparison to other open-source code models of comparable size.}
    \label{fig:mdeval}
\end{figure*}

\paragraph{MdEval} OpenCoder is also evaluated on the comprehensive multilingual code debugging benchmark MdEval~\citep{mdeval} across 18 languages. In contrast to McEval, this benchmark focuses on the assessment of code debugging, especially for language-specific bugs. Figure \ref{fig:mdeval} shows the results of the multilingual automated program repair task for OpenCoder-8B-Instruct, which comprises nearly 1.2K samples, which demonstrates that OpenCoder can effectively find the bugs and fix them compared to other open-source models of comparable size.



\section{Analysis}


\subsection{Analysis of the Deduplication Level}\label{section:code_dedup_analysis}
\begin{figure*}[h!]
    \centering
    \includegraphics[width=\textwidth]{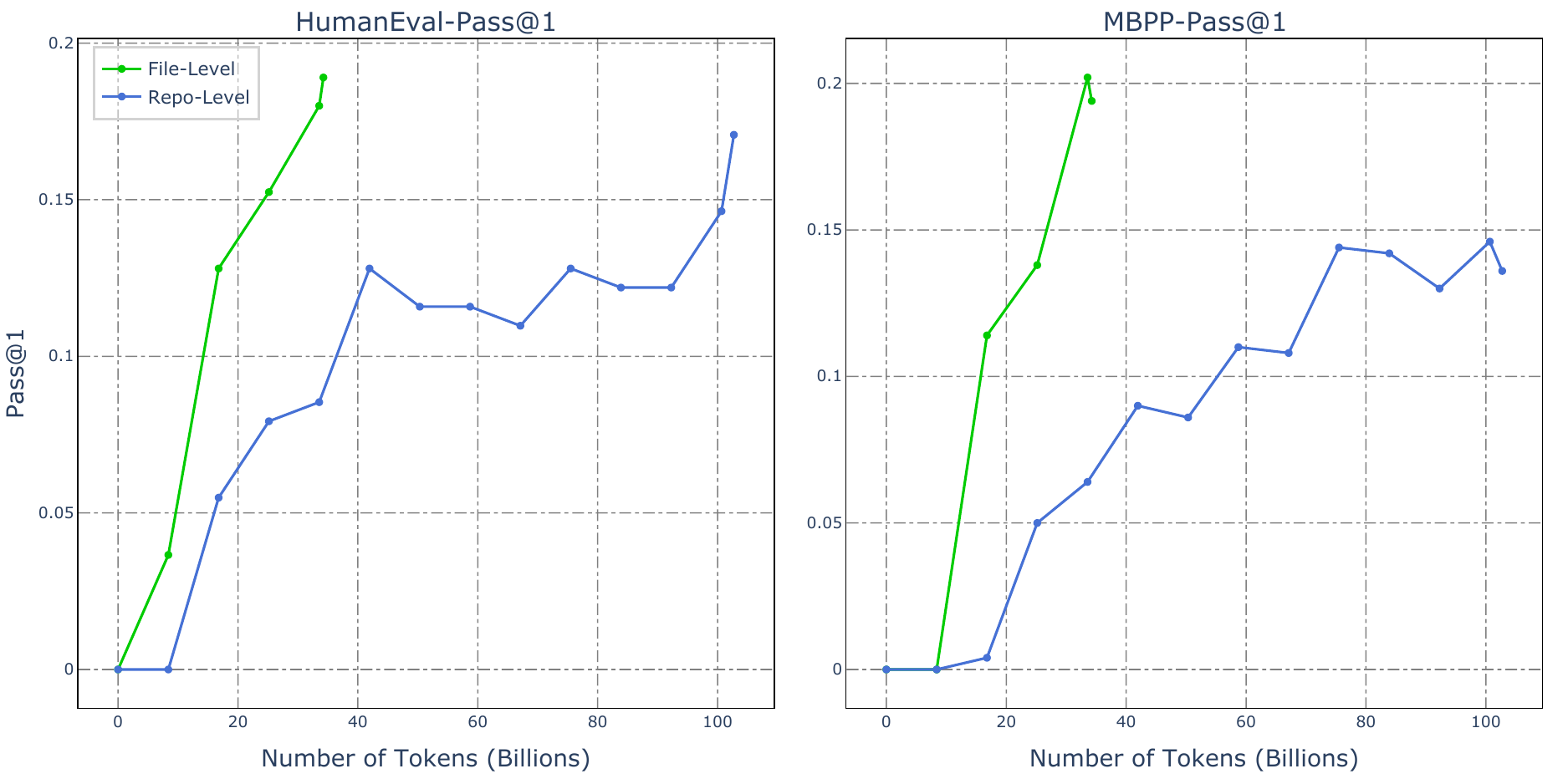} %
    \caption{Impact of using different deduplication strategies.}
    \label{fig:dedup_comparison}
\end{figure*}

Recent studies~\citep{lee2021deduplicating} have demonstrated the significant performance improvements that can be achieved by deduplicating training datasets for LLM,
where MinHash combined with LSH has emerged as the predominant method for deduplication in code training datasets~\citep{li2023starcoder, lozhkov2024starcoder, guo2024deepseek, mishra2024granite}. 
Recently, DeepSeekCoder~\citep{guo2024deepseek} claims that deduplication is performed at the repository level. 
However, we conduct extensive experiments on the Python corpus of RefineCode by performing deduplication at both the file and repository levels, respectively.Specifically, the deduplication is conducted at both the file level and repository level across the 485 million Python files available on GitHub,
respectively,
and then we train two 1.5B LLMs,
where the findings are as follows:
First, in Table~\ref{table:dedup_analyze}, the number of retained tokens at the repository level deduplication is almost three times that of the file level deduplication. 
Second, in Figure~\ref{fig:dedup_comparison}, 
we compare the downstream performance of the two datasets (i.e., HumanEval and MBPP) during pretraining and observe that the performance of file level deduplication is better than the performance of repository level deduplication a lot.
Third, for repository level deduplication, we observe that a substantial portion of 52 billion tokens exhibits complete character-level equivalence with another file.
Fourth, when conducting file-level deduplication as a post-processing step on the results of repository-level deduplication, we find that approximately 68 billion tokens (about 68.4\% of the data) could be further deduplicated. Our further investigation into chunk-level deduplication revealed no observable benefits, as detailed in the Appendix~\ref{appendix:dedup_detail}. In summary, for large-scale code datasets, performing exact deduplication followed by file-level fuzzy deduplication is an efficient and CPU-saving approach.



\begin{table}[h!]
\centering
\caption{The statistics for file level deduplication and repository level deduplication on Python code. Rows for file level and repository level represent the number of files and repositories, respectively.}
\begin{tabular}{ccccc}
  \toprule
  \textbf{Deduplication Level} & \textbf{\# Total Rows} & \textbf{\# Retained Rows} & \textbf{\# Retained Tokens} \\
  \midrule
  File level & 485,817,123 & 30,488,834 & 32.74 B (2.4\%)\\
  Repository level & 11,037,352 & 7,480,488 & 99.47 B (7.5\%)\\
  \bottomrule
\end{tabular}
\label{table:dedup_analyze}
\end{table}

\subsection{Analysis on the Importance of High-quality Data In the Annealing Phase}\label{decay_data_distribution}

\begin{figure*}[h!]
    \centering
    \includegraphics[width=\textwidth]{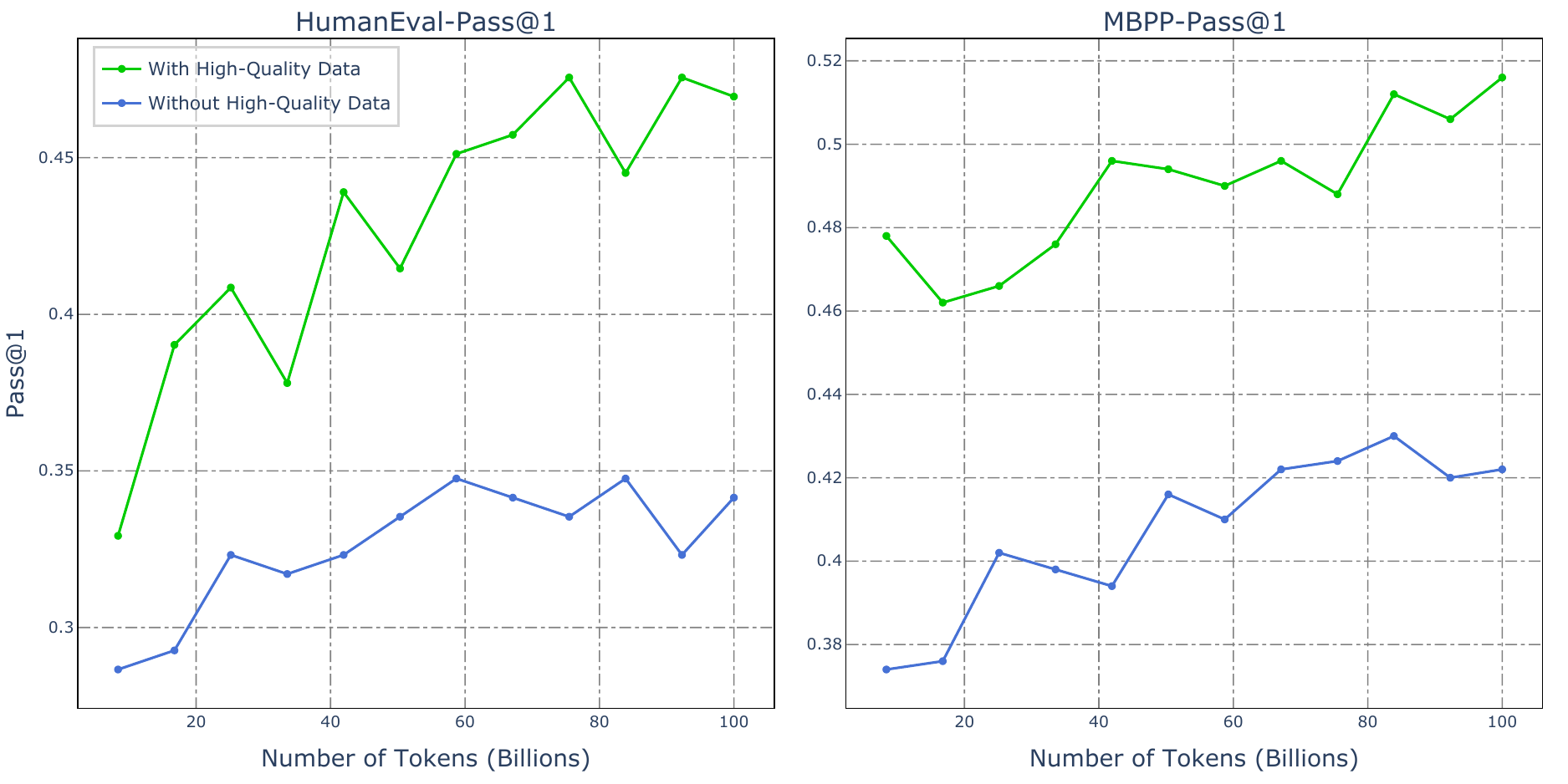} %
    \caption{Impact of using high-quality data in the annealing stage.}
    \label{fig:decay_comparison}
\end{figure*}

During the annealing phase of training, we conduct experiments by using different annealing data with different data distributions as shown in Figure~\ref{fig:decay_comparison}.
Similarly,
we still train two 1.5B LLMs,
where the first is trained by our original annealing data previously introduced and the second is trained by the data without using the high-quality data (i.e., Algorithmic Corpus and the Synthetic Data).
From Figure~\ref{fig:decay_comparison},
we observe that the performance drops a lot when the high-quality training data is removed,
which demonstrates the effectiveness of our constructed high-quality data in the annealing phase.

\subsection{Analysis on the Effect of GitHub Stars}\label{analysis_stars}
\begin{figure*}[tb]
    \centering
    \includegraphics[width=\textwidth]{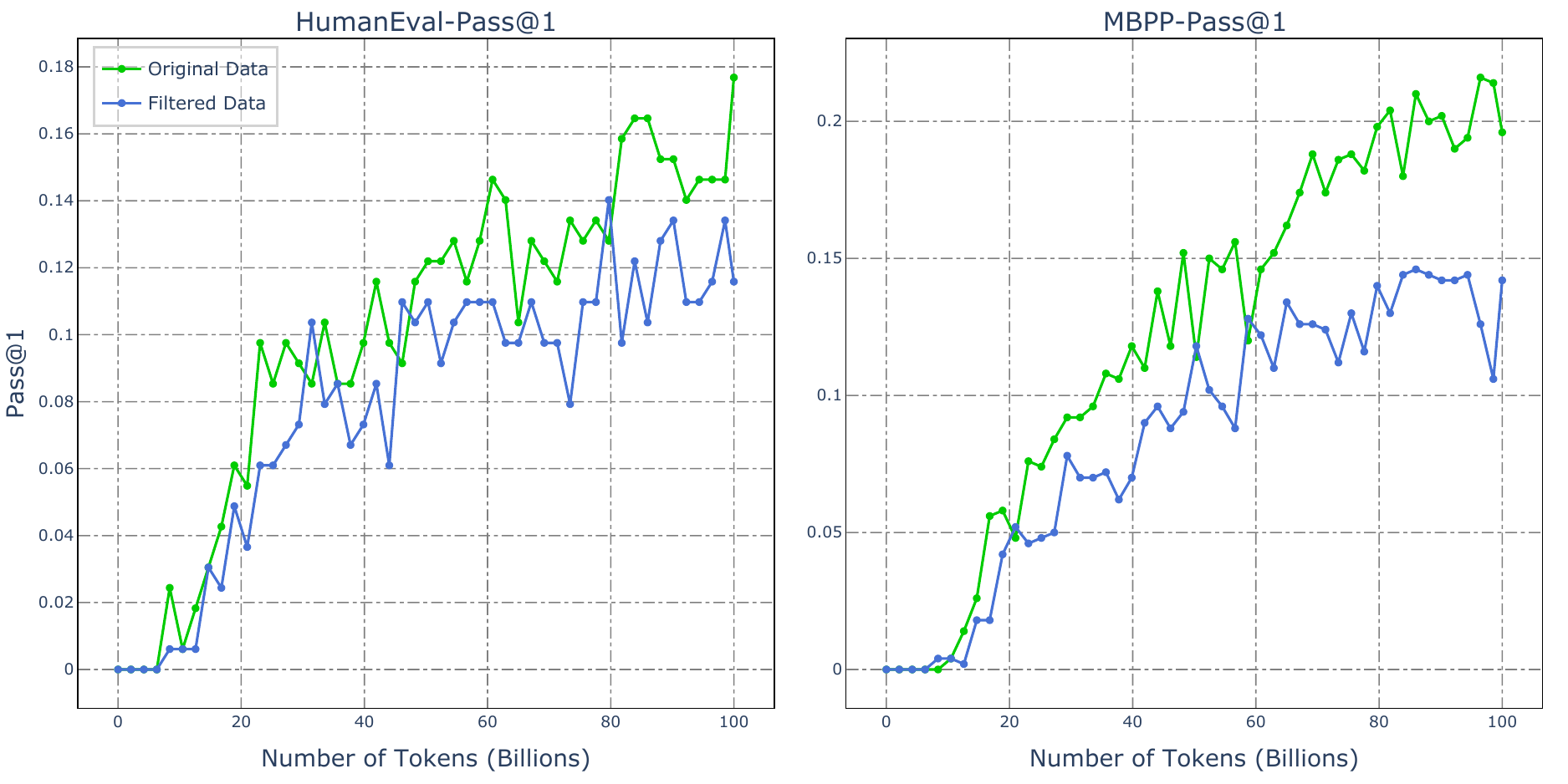} %
    \caption{Impact of star-based data filtering on model performance.}
    \label{fig:stars_comparison}
\end{figure*}

\begin{figure}[h!]
	\centering
        \includegraphics[width=\textwidth]{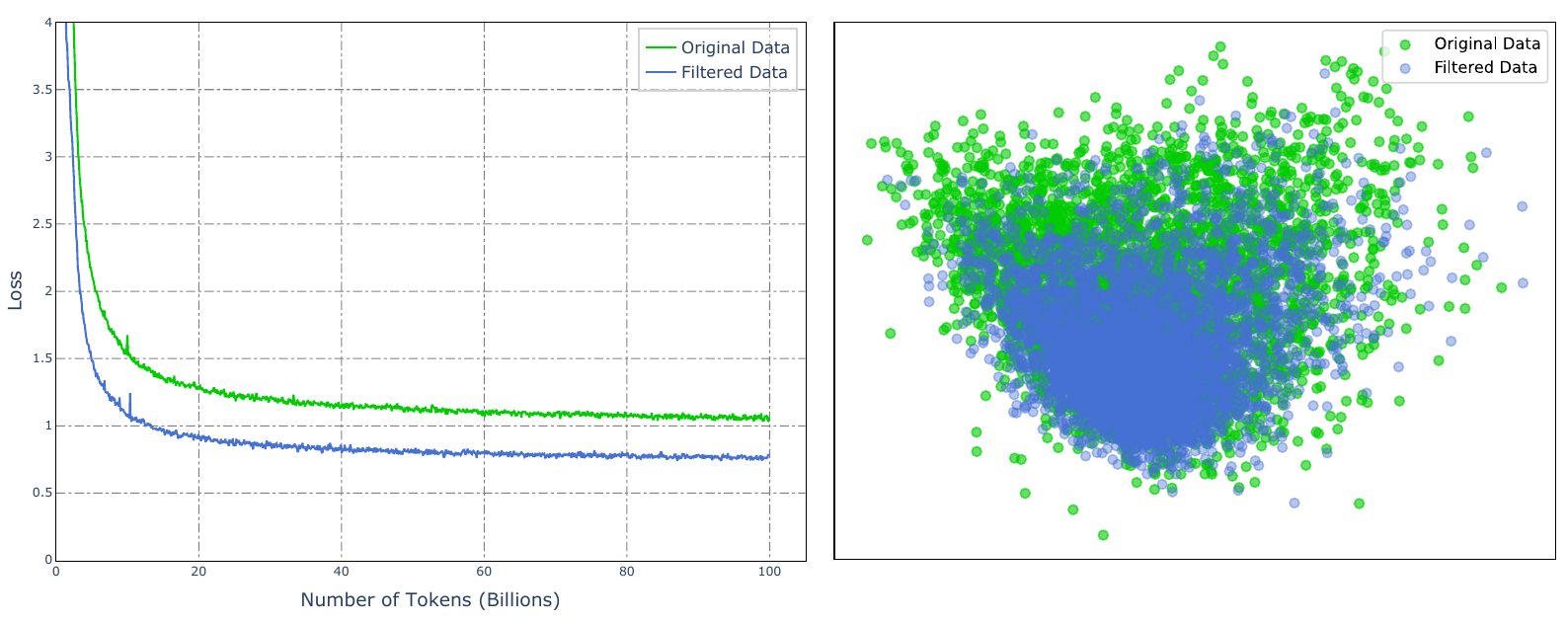}
        \caption{Left figure: Losses of using different training data with different distributions. Right figure: Visualization of the embeddings for original data and filtered data. Note that filtering based on the number of stars can reduce data diversity and result in a lower overall loss for pretraining.}
        \label{fig:loss_vis}
\end{figure}

Following SantaCoder~\citep{allal2023santacoder}, we also conduct experiments by comparing the performance trained by original code data and the filtered code data based on GitHub Stars, respectively.
Specifically,
as shown in Figure~\ref{fig:stars_comparison},
we train two 1.5B LLMs,
where one is trained original data and another is trained by data filtered by GitHub stars (stars>=5),
and we have the following findings.
First,
in Figure~\ref{fig:stars_comparison},
we observe that the LLM trained by original data is better than the LLM trained by filter data,
which is similar to the results of SantaCoder.
Second,
in Figure~\ref{fig:loss_vis},
we also provide the training losses of these two LLMs and observe that the loss of the LLM trained by filtered data is fewer than the LLM trained by original data.
For this phenomenon,
we assume that the data quality is better when using stars as the filter signal, but the diversity is relatively limited compared to the original data.
Besides, we find that this effect can be predicted from a single data distribution through visualization alone, without the need for training. As dedicated in Figure \ref{fig:loss_vis}, star filter significantly impacts the overall data distribution, compromising data diversity. Upon closer examination of the filtered data, we find that it still contains a considerable amount of well-structured, algorithmically rich code. Therefore, we argue that using stars as a filtering criterion is not an optimal choice.

\subsection{Analysis on the two-stage instruction tuning strategy}

We compared three tuning strategies for OpenCoder-1.5B-Instruct: Stage1, Stage1+Stage2, and Mix Training. Table \ref{tab:code_arena} indicates that the two-stage SFT training can bring consistent improvement in both public benchmarks and real-world scenarios. We observe that the data in Stage 1 exhibits significant diversity, though with relatively lower average quality. In contrast, the data in Stage 2 consists of high-quality, code-specific SFT  data. This two-stage SFT strategy allows for the acquisition of broad capabilities in Stage 1, followed by targeted enhancement of code-related tasks in Stage 2. Besides, similar to Chatbot Arena, we adopt the CodeArena test set covering nearly 400 human-created samples to emulate user code-related prompts in realistic environments. We use GPT-4 as the baseline and use GPT-4 to judge which LLM has a better response, where the reported results are win rate compared to the GPT-4. Table \ref{tab:code_arena} demonstrates the importance of the two-stage SFT training strategy in the algorithmic benchmarks Evalplus and the realistic benchmarks CodeArena.

\begin{table}[h]
\centering
\caption{Performance of different training strategies across benchmarks. Mix Training refers to the process of combining and shuffling the data from Stage 1 and Stage 2 for joint training.}
\begin{tabular}{l|cccccc}
\toprule
                & HE & HE+ & MBPP & MBPP+ & BigCodeBench & Code Arena \\ 
\midrule
Stage1          & 52.4 & 48.1 & 68.7 & 57.4 & 22.1 & 5.3 \\
Stage1 + Stage2 & \textbf{70.1} & \textbf{64.0} & \textbf{74.6} & \textbf{64.8} & \textbf{31.5} & \textbf{6.9} \\
Mix Training    & 55.5 & 51.2 & 52.0 & 58.7 & 23.9 & 3.8  \\
\bottomrule
\end{tabular}
\label{tab:code_arena}
\end{table}





\section{Related Work}


\noindent\textbf{Code Large Language Models.}
The remarkable progress in generative language modeling has sparked numerous studies on AI applications for software engineering~\citep{black2022gptneox, brown2020gpt3, radford2019gpt2, touvron2023llama,unicoder,mceval,mdeval}.  While proprietary models~\citep{achiam2023gpt4, chen2021evaluating, chowdhery2023palm} achieve significant performance improvements in many code-related benchmark datasets~\citep{chen2021evaluating, hendrycks2020measuring},  the inaccessible model checkpoints hinder further innovation.
In contrast, the research community has introduced several open-source models (e.g., CodeGen \citep{nijkamp2023codegen2, nijkamp2022codegen}, StarCoder \citep{li2023starcoder, lozhkov2024starcoder2}, CodeLlama \citep{roziere2023codellama} and DeepSeekCoder \citep{guo2024deepseek}), which greatly foster continued innovation in the field.

\noindent\textbf{Code Benchmarks.}
Code generation models can be leveraged to address programming challenges by interpreting and acting upon input specifications, which involves the automatic creation of programming solutions based on given problem descriptions
~\citep{athiwaratkun2022multi, austin2021program, chen2021evaluating, gu2024cruxeval, lai2023ds1000, mceval,niklas2024octopack,unicoder}.
Moreover, many benchmark datasets have been proposed to comprehensively assess code large language models, such as code retrieval~\citep{husain2019codesearchnet, lu2021codexglue}, code translation~\citep{yan2023codetransocean}, code efficiency~\citep{du2024mercury} and the challenging repository-level code completion tasks~\citep{allal2023santacoder, liu2023repobench, shrivastava2022repository, zhang2023repocoder, Deng2024R2C2CoderEA,Liu2024M2rcEvalMM,Deng2024R2C2CoderEA}.

\noindent\textbf{Open Large Language Models.}
Recently, many open-sourced LLMs have been proposed to empower the open research community and inspire a new wave of innovation. 
Specifically, 
many LLMs (e.g., LLaMA~\citep{touvron2023llama}, Mistral~\citep{Jiang2023Mistral7}, Qwen~\citep{qwen}, ChatGLM~\citep{glm2024chatglm}), pretraining datasets (e.g., RedPajama~\citep{together2023redpajama}, SlimPajama~\citep{cerebras2023slimpajama}, FineWeb~\citep{penedo2024finewebdatasetsdecantingweb}),
and chat-related datasets (e.g., WildChat~\citep{zhao2024wildchat}, LMSYS-Chat-1M~\citep{zheng2023lmsyschat1m}) are open-sourced, which greatly inspire more research innovations and accelerate the improvements of LLMs.  
Notably, several fully open LLMs have been introduced,
which provide as many details as possible to reproduce high-performance LLMs.
For example, 
in general LLMs,
OLMo~\citep{groeneveld-etal-2024-olmo}, OLMoE~\citep{olmoe},
LLM360~\citep{liu2023llm360} and MAP-Neo~\citep{zhang2024map}
are proposed.
These models release not only the final model checkpoint but also many training details (e.g., the data processing pipeline, the pretraining data, and the intermediate checkpoints).
In code LLMs,
StarCoder~\citep{allal2023santacoder} and StarCoderV2~\citep{lozhkov2024starcoder} also release high-quality code pretraining corpus.


\section{Conclusion \& Future Work}


In this paper, we present OpenCoder, an open LLM specialized in code intelligence that achieves top-tier performance. To advance research transparency and reproducibility, we release our complete training materials, including: the complete data processing pipeline, the reproducible pretraining dataset, the open code SFT dataset, rigorous experimental ablation results, detailed training protocols and intermediate checkpoints.
The performance of OpenCoder is on par with leading proprietary models, and it surpasses most previous open-source models at the both 1B+ and 6B+ parameter scale.
Furthermore, we conducted a series of ablation analyses on each phase of the code LLM training process, providing valuable insights and recommendations for future code LLM training.
We hope the release of OpenCoder can democratize access to all aspects of a top-tier code LLM, serving as both a powerful model and an open foundation to accelerate research and enable reproducible advancements in code AI.

In the future, we will continue to update our model and data consistently, aiming to improve OpenCoder's performance and expand its influence within the community. Our commitment is to ensure that OpenCoder remains at the forefront of technological advancements, providing users with the most efficient and accurate coding assistance possible. By regularly incorporating user feedback and the latest research findings, we strive to build a more robust and versatile platform that can cater to the diverse needs of developers around the world.

\bibliography{iclr2024_conference}
\bibliographystyle{iclr2024_conference}

\clearpage
\appendix
\begin{appendices}
\section{Filtering Rules}

\subsection{Design of Filtering Rules}\label{appendix:filter_design}

Designing heuristic filtering rules is inherently challenging, often requiring iterative refinement and experimentation to ultimately develop an effective set of rules. Given this complexity, in addition to providing detailed explanations of our designed rules, we will also share the general insights and methodologies we have accumulated throughout the designing process. We believe that this section will offer valuable guidance for designing heuristic filtering rules applicable to any dataset, thereby significantly enhancing the efficiency of constructing an effective data cleaning pipeline.

Heuristic rules filter data based on specific characteristics of a file, which, for each file, are ultimately expressed as a score representing the file's attribute and a corresponding threshold set by the rule. During the rule design process, we found that understanding the distribution of scores and the impact of different threshold settings on data filtering is critical to creating effective rules. Therefore, based on the approach used in RedPajama~\citep{together2023redpajama}, we decompose the heuristic filtering process into two steps: \textbf{quality signal computation} and \textbf{filtering execution}. The quality signal computation calculates the scores for all rules for each file, while the filtering execution module decides whether a file is retained based on its quality signal scores and the corresponding thresholds.

Additionally, we recommend placing the heuristic filtering process as late as possible in the overall data pipeline. Unlike other, more fixed stages of the data processing pipeline, this stage requires frequent adjustments based on the final quality of the data. Placing it later in the process allows for more precise control over the data and minimizes the need to repeat subsequent steps after this filtering module.

The specific steps for designing our heuristic filtering rules are as follows:

\begin{enumerate}

\item \textbf{Quality Signals Designing:} Based on the definition of low-quality data and the attributes of the dataset, we firstly design a series of quality signals that describe the attributes contributing to file quality.

\item \textbf{Coarse Threshold Tuning:} Referring to the definition of low-quality data and the distribution of quality signal scores, we roughly set filtering thresholds for all rules at once. We then apply the filters to obtain an initial version of the filtered dataset.

\item \textbf{Fine-grained Threshold Tuning:} For each rule, we focus on the data that was exclusively affected by that specific rule, meaning it did not trigger other filters. This part of the data is directly influenced by the current rule, so we can examine whether the retention or removal of this data under different threshold settings aligns with the intended purpose of the rule. If a rule is effective in improving data quality based on its target attribute, we select the optimal threshold; otherwise, the rule is discarded. After evaluating each rule, we apply the filters again to obtain a more refined filtered dataset.

\item \textbf{Data Quality Inspection:} We then assess whether the filtered dataset meets our expectations for the quality of pretraining data. In addition to traditional manual inspection, we introduce a perplexity (PPL)-based method for data quality evaluation. Specifically, we randomly sample a set of data from the filtered dataset and use a high-performing LLM to compute the PPL on these samples. We then examine the top-N and bottom-N samples based on PPL. Generally, extremely low PPL suggests that the data is overly simplistic, containing limited valuable knowledge, while extremely high PPL indicates that the data may lack learnable patterns. Both of them are advisable to be filtered out. We closely inspect both sets of samples and, based on their characteristics, decide whether to add new rules or adjust existing thresholds. This process can be repeated until the dataset reaches the desired quality.

\end{enumerate}

\subsection{Examples of Filtering Rules}\label{appendix:filter_rules}
We elaborate several representative examples about general code filtering rules in Table~\ref{table:code_code_rules} and language-specific filtering rules in Table \ref{table:specific_code_rules} and explain their rationale. It is essential to note that for general code filtering rules, the threshold values may be slightly adjusted depending on the programming language of the file. For specific threshold values, please refer to our implementation details of the data processing pipeline. 

\newcolumntype{Y}{>{\RaggedRight\arraybackslash}X}
\newcolumntype{Z}{>{\RaggedRight\arraybackslash}X}

\begin{table*}[h!]
\centering
\caption{Examples of general code filtering rules.}
\begin{tabularx}{\textwidth}{>{\hsize=0.38\hsize}Y >{\hsize=0.32\hsize}Z >{\hsize=0.3\hsize}X}
    \toprule
    Description & Explanation & Filtering Quota \\
    \midrule
    The proportion of lines in strings with a word count exceeding. & Files with too many long strings indicate a lack of code logic. & score > 0.2 \\
    \midrule
    The proportion of characters in words from strings with a character count exceeding 20. & String variables containing long sequences of characters are often indicative of meaningless content such as base64 data, Hash encoding, url, etc.  & score > 0.4 \\
    \midrule
    The proportion of hexadecimal characters. & Files with two many hexadecimal characters indicate a lack of code logic. & score > 0.4 \\
    \midrule
    The proportion of lines like "you code here", "TODO" or "FIXME". & We found that these elements tend to be excessively repeated in the dataset, which increases the likelihood that the model, during code completion, will output placeholders like the ones mentioned above instead of generating actual code. & score > 0.01 \\
    \midrule
    The proportion of lines containing an "assert" statement. & Files containing a large number of 'assert' statements are often test files, which tend to have relatively simple and repetitive code patterns. & score > 0.4 \\
    \bottomrule
\end{tabularx}
\label{table:code_code_rules}
\end{table*}

\begin{table*}[h!]
\centering
\caption{Examples of python-specific filtering rules.}
\begin{tabularx}{\textwidth}{>{\hsize=0.38\hsize}Y >{\hsize=0.32\hsize}Z >{\hsize=0.3\hsize}X}
    \toprule
    Description & Explanation & Filtering Quota \\
    \midrule
    The proportion of the number of python functions to the total number of lines. & A higher number of Python functions in a file may indicate that the functions are overly simple, with limited code logic, or have a bad code format. & score > 0.2 \\
    \midrule
    Whether the file can be parsed into an python abstract syntax tree (AST). & Files that cannot be parsed into an AST contain syntax errors and should be filtered out. & score == False \\
    \midrule
    The proportion of lines that are "import" statements. & A file with exceeding prportion of "import" statements indicates to have sparse code logic. & score > 0.3 \\
    \bottomrule
\end{tabularx}
\label{table:specific_code_rules}
\end{table*}

\end{appendices}
\clearpage
\appendix
\begin{appendices}

\section{Analysis on Chunk-level Deduplication}\label{appendix:dedup_detail}

During pretraining, data is first randomly concatenated and segmented into chunks of context length, followed by full-attention computation within each chunk. We further explored chunk-level deduplication. Specifically, the pretraining data was randomly concatenated and segmented into chunks of 4096 tokens, followed by MinHash and LSH deduplication on these chunks. Additionally, we applied chunk-level deduplication after file-level and repo-level deduplication. 

\begin{table}[h!]
\centering
\caption{Comparison of deduplication strategies on Python data. At the File level, "Lines" refers to the number of lines in individual files; at the Repo level, it indicates the line count of aggregated strings; Note that for all deduplication strategies involving the Chunk level, "Lines" specifically refers to 4096-token chunks.}
\begin{tabular}{l|rrr}
\toprule
                & \# Total Lines & \# Retained Lines & \# Retained Tokens \\
    \midrule
    Chunk-level & 333,007,812 & 79,272,460 & 324.70 B  \\
    File-level & 485,817,123 & 30,488,834 & 32.74 B  \\
    File-level + Chunk-level & 333,007,812 & 7,993,164 & 32.70 B  \\
    Repo-level & 11,037,352 & 7,480,488 & 99.47 B \\
    Repo-level + Chunk-level & 333,007,812 & 17,675,781 & 72.40 B \\
\bottomrule
\end{tabular}
\label{tab:dedup_analyze}
\end{table}

\begin{figure*}[h!]
\centering
\includegraphics[width=1.0\textwidth]{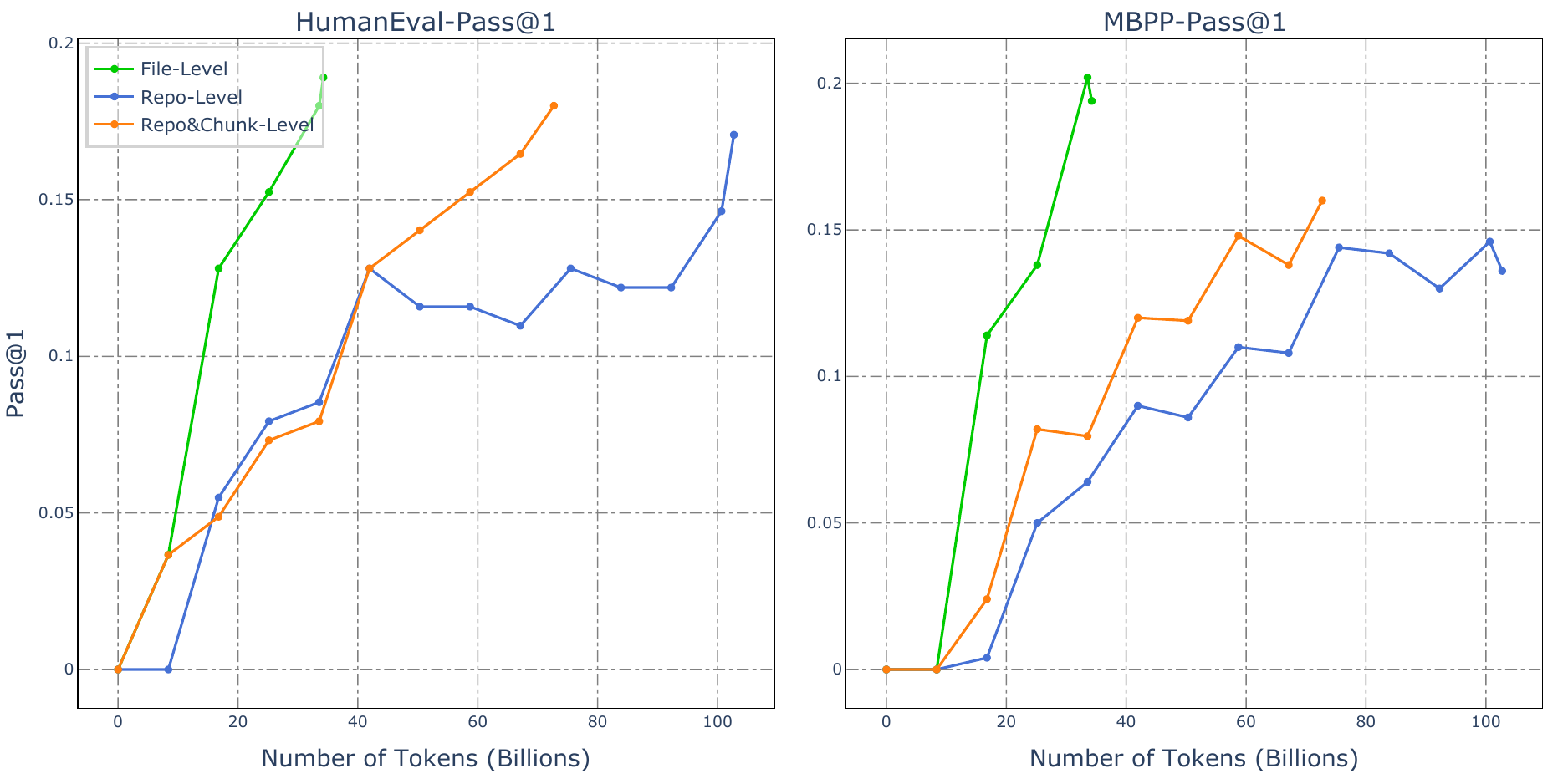} %
\caption{Comparison of Pass@1 performance on HumanEval \& MBPP for different dedup strategies (File-Level, Repo-Level, and Repo-level + Chunk-Level) across RefineCode Python corpus.}
\label{fig:chunk_dedup_analyze_2}
\end{figure*}

From the results in table \ref{tab:dedup_analyze}, We observe that chunk-level deduplication alone was even less effective than repo-level deduplication, and applying chunk-level deduplication after file-level removed only an additional 0.04B of data. This indicates that chunk-level deduplication is not an effective approach. We pre-trained three 1.5B models on the data retained under file-level, repo-level, and repo-level + chunk-level deduplication strategies. The benchmark results are shown in Figure \ref{fig:chunk_dedup_analyze_2}. It is evident that file-level deduplication achieves the highest training efficiency, while repo-level + chunk-level deduplication outperforms repo-level alone. We attribute the superior performance of file-level deduplication to its higher degree of data removal.
Overall, we conclude that file-level deduplication is the most suitable method for GitHub data.

\end{appendices}

\appendix
\begin{appendices}
\section{Extra Data Processing}\label{appendix:extra_data}
\subsection{Chinese Code-Like Domains Annotation}
The manual annotation of the URLs of the website is presented as shown in the table \ref{table:cc_annotate}. For future new CC datasets, we can sample pages in these domains as initial seed corpus. 

\begin{table}[ht]
\centering
\caption{We manually annotate code-like and math-like Chinese domains, utilizing the '\%' symbol as a wildcard in our pattern matching. For example, the URL 'https://my.oschina.net/u/4/blog/11' is matched by the pattern '\%my.oschina.net\%blog\%'.}
\begin{tabular}{lcc}
\toprule
Domain & Prefix & Tag \\
\midrule
cloud.tencent.com & \%cloud.tencent.com/developer/article\% & Code \\
cloud.tencent.com & \%cloud.tencent.com/ask\% & Code \\
cloud.tencent.com & \%cloud.tencent.com/developer/information\% & Code \\
cloud.tencent.com & \%cloud.tencent.com/document\% & Code \\
my.oschina.net & \%my.oschina.net\%blog\% & Code \\
ask.csdn.net & \%ask.csdn.net/questions\% & Code \\
www.cnblogs.com & \%www.cnblogs.com\% & Code \\
forum.ubuntu.org.cn & \%forum.ubuntu.org.cn\% & Code \\
q.cnblogs.com & \%q.cnblogs.com/q\% & Code \\
segmentfault.com & \%segmentfault.com/q\% & Code \\
segmentfault.com & \%segmentfault.com/a\% & Code \\
woshipm.com & \%woshipm.com/data-analysis\% & Code \\
zgserver.com & \%zgserver.com/server\% & Code \\
zgserver.com & \%zgserver.com/linux\% & Code \\
zgserver.com & \%zgserver.com/ubuntu\% & Code \\
juejin.cn & \%juejin.cn/post\% & Code \\
jiqizhixin.com & \%jiqizhixin.com/articles\% & Code \\
help.aliyun.com & \%help.aliyun.com/zh\% & Code \\
jyeoo.com & \%jyeoo.com\% & Math \\
www.haihongyuan.com & \%haihongyuan.com\%shuxue\% & Math \\
www.03964.com & \%www.03964.com\% & Math \\
www.nbhkdz.com & \%www.nbhkdz.com\% & Math \\
9512.net & \%9512.net\% & Math \\
lanxicy.com & \%lanxicy.com\% & Math \\
bbs.emath.ac.cn & \%bbs.emath.ac.cn\% & Math \\
math.pro & \%math.pro\% & Math \\
mathschina.com & \%mathschina.com\% & Math \\
shuxue.chazidian.com & \%shuxue.chazidian.com\% & Math \\
shuxue.ht88.com & \%shuxue.ht88.com\% & Math \\
\bottomrule
\end{tabular}
\label{table:cc_annotate}
\end{table}

\subsection{Code-Related Data from Github Text Files}
Github Text files primarily consist of content written in natural languages, which includes abundant code-related knowledge. However, we observed that a substantial portion of the dataset is unrelated to code, which is detrimental to the model's ability to learn code-related knowledge. Therefore, we employed the following strategies to extract and retain the code-relevant portions before our filtering module. Firstly, following the strategy used in starcoder~\citep{li2023starcoder}, we retained the files with "requirement" in the lowercased filename, or if the filename without the extension is one of {"readme", "notes", "todo", "description", "cmakelists"}, in order to ensure that only text files pertinent to coding contexts are preserved. This strategy recalled 3\% volume of the whole text part. Additionally, we trained a fasttext model to recall code-related text files and recalled extra 7\% file volume from the original text data. 

\subsection{Jupyter Notebooks}
Our Jupyter notebook data is sourced from GitHub and Meta Kaggle code~\citep{jim_plotts_megan_risdal_2023}. We converted this type of data into the \textit{Jupyter-structured} format used in StarCoder~\citep{li2023starcoder}, which consists of a triplet of consecutive markdown, code, and code execution results. However, we discarded the \textit{Jupyter-script} format mentioned in StarCoder. Because the code files generated from Jupyter notebook conversions tend to have poor overall code writing standards, and the content in \textit{Jupyter-script} and \textit{Jupyter-structured} formats is highly redundant, making it sufficient to retain only one format.

\end{appendices}
\appendix
\begin{appendices}
\section{Comparison of RefineCode with The Stack Series}\label{appendix:comparison_of_datasets}
Table \ref{table:corpus_comparison}  compares RefineCode with two versions of The Stack. RefineCode not only includes more tokens (960 billion) but also incorporates over 130 rules, significantly more than the 15 rules used in previous versions. Additionally, RefineCode leverages 75 billion web data tokens and introduces language-specific (LS) rules, providing more precise and fine-tuned handling across a wide range of programming languages. 

\begin{table}[h!]
\centering
\caption{The Comparison of training data between \textbf{RefineCode} and series of The Stack. ``LS'' denotes ``Language Specific''.}
\begin{tabular}{c|ccccc}
  \toprule
  & \textbf{\# Tokens} & \textbf{\# Languages} & \textbf{\# Web Data Tokens} & \textbf{\# Rules} & \textbf{LS Rules}\\
  \midrule
  The Stack v1 & 200 B & 88 & \textbackslash & \textasciitilde 15 & \xmark \\
  The Stack v2 & 900 B & 619 & \textasciitilde 30 B & \textasciitilde 15 & \xmark \\
  \textbf{RefineCode} & 960 B & 607 & \textasciitilde 75 B & \textasciitilde 130 & \cmark \\
  \bottomrule
  \end{tabular}
\label{table:corpus_comparison}
\end{table}

\end{appendices}
\appendix
\begin{appendices}

\section{Programming Languages Categories}\label{appendix:program_langs}

\subsection{Included Programming Languages}
Included programming languages can be categoried into three classes: code, data and text. Among them, the "code" category represents files rich in code logic, while the "data" category primarily consists of files with structured data, and the "text" category refers to files dominated by natural language content. The threshold settings for the filtering rules vary slightly depending on the data type.

\paragraph{Code(470 types):}1C Enterprise, 4D, ABAP, ABAP CDS, AIDL, AL, AMPL, ANTLR, API Blueprint, APL, ASL, ASP.NET, ATS, ActionScript, Ada, Agda, Alloy, Alpine Abuild, AngelScript, Apex, Apollo Guidance Computer, AppleScript, Arc, AspectJ, Assembly, Astro, Asymptote, Augeas, AutoHotkey, AutoIt, Awk, BASIC, BQN, Ballerina, Batchfile, Beef, Befunge, Berry, Bikeshed, Bison, BitBake, Blade, BlitzBasic, BlitzMax, Bluespec, Boo, Boogie, Brainfuck, Brightscript, C, C\#, C++, C2hs Haskell, CAP CDS, CLIPS, CMake, COBOL, CUE, Cadence, Cairo, CameLIGO, Cap'n Proto, Ceylon, Chapel, Charity, ChucK, Circom, Cirru, Clarion, Clarity, Classic ASP, Clean, Click, Clojure, Closure Templates, CodeQL, CoffeeScript, ColdFusion, ColdFusion CFC, Common Lisp, Common Workflow Language, Component Pascal, Coq, Crystal, Csound, Csound Document, Csound Score, Cuda, Curry, Cycript, Cypher, Cython, D, D2, DIGITAL Command Language, DM, Dafny, Dart, DataWeave, Dhall, Diff, Dockerfile, Dogescript, Dylan, E, ECL, EJS, EQ, Earthly, Edge, EdgeQL, Elixir, Elm, Elvish, Emacs Lisp, EmberScript, Erlang, F\#, F*, FIRRTL, FLUX, Factor, Fancy, Fantom, Faust, Fennel, Filebench WML, Fluent, Forth, Fortran, Fortran Free Form, FreeBasic, Futhark, GAML, GAMS, GAP, GDB, GLSL, GSC, Game Maker Language, Genero 4gl, Genero per, Genshi, Gentoo Ebuild, Gentoo Eclass, Gherkin, Gleam, Glimmer JS, Glyph, Go, Golo, Gosu, Grace, Grammatical Framework, Groovy, Groovy Server Pages, HCL, HLSL, HTML, HTML+ECR, HTML+EEX, HTML+ERB, HTML+PHP, HTML+Razor, Hack, Haml, Handlebars, Harbour, Haskell, Haxe, HiveQL, HolyC, Hy, IDL, IGOR Pro, Idris, ImageJ Macro, Imba, Inform 7, Ink, Inno Setup, Io, Ioke, Isabelle, Isabelle ROOT, J, JCL, JFlex, JSONiq, Janet, Jasmin, Java, Java Server Pages, JavaScript, JetBrains MPS, Jinja, Jison, Jison Lex, Jolie, Jsonnet, Julia, Just, KRL, Kaitai Struct, KakouneScript, KerboScript, Kit, Kotlin, LFE, LLVM, LOLCODE, LSL, LabVIEW, Latte, Lean, Less, Lex, LigoLANG, LilyPond, Limbo, Liquid, Literate Agda, Literate CoffeeScript, Literate Haskell, LiveScript, Logos, Logtalk, LookML, Lua, Luau, M, M4, M4Sugar, MATLAB, MAXScript, MLIR, MQL4, MQL5, MTML, MUF, Macaulay2, Makefile, Mako, Marko, Mask, Mathematica, Mercury, Mermaid, Meson, Metal, MiniD, Mint, Mirah, Modelica, Modula-3, Module Management System, Mojo, Monkey, MoonScript, Motorola 68K Assembly, Move, Mustache, Myghty, NASL, NSIS, NWScript, Nearley, Nemerle, NetLinx, NetLogo, Nextflow, Nim, Nit, Nix, Nu, NumPy, Nunjucks, OCaml, Oberon, Objective-C++, Objective-J, Omgrofl, Opa, Opal, Open Policy Agent, OpenCL, OpenQASM, OpenSCAD, Ox, Oxygene, Oz, P4, PDDL, PEG.js, PHP, PLSQL, PLpgSQL, Pact, Pan, Papyrus, Parrot, Parrot Assembly, Parrot Internal Representation, Pascal, Pawn, Pep8, Perl, PigLatin, Pike, PogoScript, Polar, Pony, Portugol, PowerBuilder, PowerShell, Praat, Processing, Procfile, Prolog, Promela, Propeller Spin, Pug, Puppet, PureScript, Prover9, Pyret, Python, Q\#, QML, QMake, Qt Script, Quake, R, RAML, REALbasic, REXX, RPGLE, RUNOFF, Racket, Ragel, Raku, Rascal, ReScript, Reason, ReasonLIGO, Rebol, Red, Redcode, RenderScript, Ring, Riot, RobotFramework, Roc, Rouge, Ruby, Rust, SAS, SMT, SQF, SQL, Sage, SaltStack, Sass, Scala, Scaml, Scenic, Scheme, Scilab, Self, Shell, ShellSession, Shen, Sieve, Singularity, Slash, Slim, Slint, SmPL, Smali, Smalltalk, Smarty, Smithy, Snakemake, SourcePawn, Squirrel, Stan, Standard ML, Starlark, Stata, Stylus, SugarSS, Svelte, Sway, Swift, SystemVerilog, TI Program, TL-Verilog, TLA, TSX, TXL, Talon, Tcl, Tcsh, Tea, Terraform Template, Thrift, Toit, Turing, Twig, TypeScript, Typst, Unified Parallel C, Uno, UnrealScript, UrWeb, V, VBA, VBScript, VCL, VHDL, Vala, Velocity Template Language, Verilog, Vim Script, Vim Snippet, Visual Basic .NET, Visual Basic 6.0, Volt, Vue, Vyper, WDL, WGSL, WebAssembly, WebIDL, Whiley, Witcher Script, Wollok, Wren, X10, XC, XProc, XQuery, XS, XSLT, Xojo, Xonsh, Xtend, YARA, YASnippet, Yacc, Yul, ZAP, ZIL, Zeek, ZenScript, Zephir, Zig, Zimpl, eC, fish, hoon, kvlang, mIRC Script, mcfunction, mupad, nesC, ooc, templ, wisp, xBase

\paragraph{Data(115 types):}ABNF, ASN.1, Adobe Font Metrics, Altium Designer, Ant Build System, ApacheConf, Avro IDL, BibTeX, Browserslist, CIL, CODEOWNERS, CSON, CSS, Cabal Config, Caddyfile, CartoCSS, Cloud Firestore Security Rules, CoNLL-U, DNS Zone, Darcs Patch, Debian Package Control File, Dotenv, EBNF, Eagle, Easybuild, Ecere Projects, EditorConfig, Edje Data Collection, FIGlet Font, Formatted, GEDCOM, GN, Gemfile.lock, Gerber Image, Git Attributes, Git Config, Glyph Bitmap Distribution Format, Go Checksums, Go Module, Go Workspace, Godot Resource, Gradle, Gradle Kotlin DSL, GraphQL, Graphviz (DOT), HAProxy, HOCON, HTTP, HXML, INI, Ignore List, JAR Manifest, JSON, JSON with Comments, Jest Snapshot, Kusto, Lark, Linker Script, Maven POM, NEON, NL, NPM Config, Nginx, Ninja, ObjDump, Object Data Instance Notation, OpenStep Property List, OpenType Feature File, Option List, PlantUML, PostCSS, Prisma, Protocol Buffer, Protocol Buffer Text Format, Python traceback, RBS, RON, Readline Config, Record Jar, Redirect Rules, Regular Expression, SCSS, SELinux Policy, SPARQL, SSH Config, STAR, STON, ShellCheck Config, Simple File Verification, Soong, Spline Font Database, TOML, TextMate Properties, Turtle, Type Language, Valve Data Format, Wavefront Material, Web Ontology Language, WebAssembly Interface Type, Wget Config, Windows Registry Entries, X BitMap, X Font Directory Index, XCompose, XML, XML Property List, XPages, YAML, YANG, cURL Config, crontab, desktop, dircolors, edn, nanorc

\paragraph{Text(22 types):}AsciiDoc, Creole, Gemini, Gettext Catalog, MDX, Markdown, Muse, Org, Pod, Pod 6, RDoc, RMarkdown, Rich Text Format, Roff, SRecode Template, Sweave, TeX, Texinfo, Text, Textile, Wikitext, reStructuredText

\subsection{Excluded Programming Languages}
2-Dimensional Array, AGS Script, Adblock Filter List, Bicep, COLLADA, CSV, Checksums, DirectX 3D File, E-mail, G-code, Git Revision List, Gnuplot, IRC log, KiCad Layout, KiCad Legacy Layout, KiCad Schematic, Lasso, Linux Kernel Module, Max, Microsoft Developer Studio Project, Microsoft Visual Studio Solution, POV-Ray SDL, Pic, Pickle, PostScript, Public Key, Pure Data, PureBasic, Raw token data, Roff Manpage, STL, SVG, SubRip Text, TSV, Unity3D Asset, Wavefront Object, WebVTT, X PixMap, robots.txt

\end{appendices}
\appendix
\begin{appendices}

\section{Raw Code Data Composition}\label{appendix:data_composition}

Figure~\ref{table:cc_annotate2} shows the composition of raw code data for top 85 programming languages in the \textbf{RefineCode} dataset, both after deduplication and filtering process. It can be observed that, after filtering, the proportion of data for different programming languages has shifted significantly, with a notable increase in the representation of commonly used programming languages.

\begin{table}[h!]
\centering
\caption{Overview of the data composition of in \textbf{RefineCode}. The items in the table are sorted in descending order according to the file volume after filtering.}
\begin{tabular}{l|rrr|rrr}
\toprule
\multirow{2}{*}{\textbf{Language}} & \multicolumn{3}{c|}{\textbf{After deduplication}} & \multicolumn{3}{c}{\textbf{After filtering}} \\
    & \textbf{\# Files} & \textbf{Vol(GB)} & \textbf{Ratio(\%)} & \textbf{\# Files} & \textbf{Vol(GB)} & \textbf{Ratio(\%)} \\
\midrule
html & 141,081,897 & 3,175.4 & 8.56 & 45,100,466 & 582.4 & 18.08 \\
java & 215,177,833 & 706.8 & 1.90 & 124,751,295 & 474.3 & 14.72 \\
python & 109,725,362 & 493.3 & 1.33 & 58,640,346 & 271.1 & 8.41 \\
csharp & 88,825,202 & 364.2 & 0.98 & 57,910,485 & 232.4 & 7.21 \\
javascript & 190,670,421 & 1,925.0 & 5.19 & 69,579,517 & 226.9 & 7.04 \\
php & 84,378,361 & 374.4 & 1.01 & 60,089,397 & 222.7 & 6.91 \\
cpp & 51,362,503 & 375.2 & 1.01 & 38,037,406 & 176.9 & 5.49 \\
go & 35,649,865 & 301.1 & 0.81 & 26,723,829 & 153.7 & 4.77 \\
typescript & 40,211,985 & 287.4 & 0.77 & 20,621,755 & 140.4 & 4.35 \\
ruby & 15,735,042 & 244.5 & 0.66 & 8,285,561 & 122.7 & 3.81 \\
perl & 16,354,543 & 121.7 & 0.33 & 9,532,620 & 65.6 & 2.04 \\
rust & 10,605,421 & 63.6 & 0.17 & 6,086,150 & 39.9 & 1.24 \\
r & 6,132,978 & 92.5 & 0.25 & 4,803,109 & 34.7 & 1.08 \\
swift & 4,238,754 & 47.9 & 0.13 & 2,938,498 & 31.8 & 0.99 \\
kotlin & 4,493,548 & 56.4 & 0.15 & 3,123,156 & 29.8 & 0.94 \\
dart & 4,087,329 & 33.0 & 0.09 & 2,161,462 & 18.5 & 0.57 \\
java-pages & 6,174,654 & 31.0 & 0.08 & 4,145,336 & 15.4 & 0.48 \\
css & 39,822,744 & 241.5 & 0.65 & 15,771,061 & 15.3 & 0.47 \\
lua & 4,027,221 & 116.0 & 0.31 & 2,538,234 & 14.4 & 0.45 \\
xml & 61,171,289 & 1,934.2 & 5.21 & 3,173,128 & 12.8 & 0.40 \\
scala & 5,897,567 & 19.7 & 0.05 & 4,204,979 & 11.7 & 0.36 \\
shell & 12,054,632 & 23.0 & 0.06 & 6,043,070 & 11.2 & 0.35 \\
pascal & 1,306,130 & 27.8 & 0.07 & 960,497 & 9.5 & 0.29 \\
fortran & 2,274,663 & 39.7 & 0.10 & 1,218,491 & 8.6 & 0.27 \\
perl6 & 1,943,430 & 16.4 & 0.04 & 1,034,748 & 8.6 & 0.27 \\
rmarkdown & 1,317,760 & 14.0 & 0.04 & 827,951 & 7.9 & 0.25 \\
html+erb & 7,618,377 & 11.4 & 0.03 & 4,452,355 & 7.8 & 0.24 \\
smali & 3,457,531 & 37.9 & 0.10 & 1,408,274 & 7.4 & 0.23 \\
scss & 18,061,278 & 35.6 & 0.10 & 7,705,822 & 7.4 & 0.23 \\
gettext catalog & 1,100,044 & 51.3 & 0.14 & 442,385 & 6.3 & 0.19 \\
haskell & 1,746,444 & 24.0 & 0.06 & 1,218,491 & 6.8 & 0.27 \\
tcl & 253,345 & 4.2 & 0.01 & 136,171 & 1.0 & 0.03 \\
gradle & 2,431,985 & 2.9 & 0.01 & 724,609 & 1.0 & 0.03 \\
scheme & 357,909 & 4.7 & 0.01 & 201,170 & 1.0 & 0.03 \\
qml & 354,756 & 1.8 & 0.01 & 252,621 & 1.0 & 0.03 \\
mdx & 795,525 & 6.4 & 0.17 & 222,013 & 1.0 & 0.03 \\
classic asp & 220,344 & 2.8 & 0.08 & 141,236 & 0.9 & 0.03 \\
xbase & 192,780 & 2.5 & 0.07 & 80,396 & 0.9 & 0.03 \\
ini & 7,232,136 & 19.1 & 0.05 & 1,517,099 & 1.3 & 0.04 \\
objective-c++ & 197,416 & 2.4 & 0.01 & 149,223 & 1.3 & 0.04 \\
motorola68k & 1,066,095 & 26.5 & 0.07 & 220,218 & 1.2 & 0.04 \\
gap & 752,261 & 2.6 & 0.01 & 510,420 & 1.2 & 0.04 \\
\bottomrule
\end{tabular}
\label{table:cc_annotate2}
\end{table}

\end{appendices}
\appendix
\begin{appendices}

\section{Prompts For SFT Synthetic Data}\label{appendix:sfr_prompts}


\begin{tcolorbox}[title=Prompt for Educational Instruction Synthesis, label={fig:test_generation_prompt}]
You are a teaching assistant helping to create a Python programming task from a given code snippet. You must provide the best response to the Python programming task, including reasoning thought, reference solutions, explanation of test cases, and test code.\\

\textbf{[Code Snippet]}\\
\{Code\}\\

Your response must have these parts:\\

\textbf{[Task]}\\
\{Create an independent and detailed Python programming task\}\\

\textbf{[Analysis]}\\
\{Analyze the task and reason about the given task step by step\}\\

\textbf{[Solution]}\\
\{Write a high-quality reference solution in a self-contained script that solves the task\}\\

\textbf{[Test]}\\
\{Provide ten assert statements to check the correctness of your solution\}\\

\end{tcolorbox}

\begin{tcolorbox}[title=Prompt for Package-related Instruction Synthesis]

You are exceptionally skilled at crafting high-educational level problems and offering precise solutions. Please gain inspiration from the following code snippet to create a high-quality programming problem, which is beneficial for learning the use of corresponding libraries. Present your output in two distinct sections: [Problem Description] and [Solution]. \\

\textbf{[Code Snippet]} \\
\{Code\} \\

\textbf{[Library Api Requirements]} \\
\{Api\ Requirements\} \\

\textbf{[Library Api Doc]} \\
\{Api\ Doc\} \\

Guidelines for each section: \\
1. [Problem Description]: This should be **completely self-contained**, providing all the contextual information one needs 
to understand and solve the problem. Assume common programming knowledge, but ensure that any specific context, variables, 
or code snippets pertinent to this problem are explicitly included. This problem should be **educational for learning the 
provided Library api, and please explicitly request the use of the relevant package in the question. This question should only 
concern the writing of **one function**, and you need to be clear about the function name and role of this function. \\

2. [Solution]: Offer a comprehensive, **correct** solution that addresses the [Problem Description] you provided. This solution 
should follow the standard of corresponding Library Api doc. Please ensure that the Solution only involves answering the Problem, 
**without addressing the requirements I provided!** Please provide essential explanation abouth this solution, especially the use
of requiremed Library Api.
    
\end{tcolorbox}


\begin{tcolorbox}[title=Prompt for Large-scale Diverse Instruction Synthesis, label={fig:question_generation_prompt}]
You are an expert in designing high-quality programming questions based on the given text.
\\

\textbf{[Guidelines]}\\
- You can draw inspiration from the given text to create the programming questions. \\
- The created question should be a self-contained question, which does not depend on any external context. \\
- The created response must contain the complete code snippet.
\\

\textbf{[Given Text]}\\
\{Given Text\}\\

\textbf{[Created Question]}\\
\{Created Question\}\\

\end{tcolorbox}

\end{appendices}

\end{document}